\documentclass[10pt,twocolumn,letterpaper]{article}

\usepackage[pagenumbers]{cvpr}   

\author{%
  \textbf{Changin Choi}$^{1,6}$, \textbf{Jungwon Park}$^{2,5}$,
  \textbf{Jungmin Ko}$^{1}$, \textbf{Jimyeong Kim}$^{3}$,
  \textbf{Wonjong Rhee}$^{1,3,4}$\\
  $^{1}$IPAI, $^{2}$RICS, $^{3}$AIIS, $^{4}$Department of Intelligence and Information,\\
  Seoul National University\\
  $^{5}$Daegu Gyeongbuk Institute of Science and Technology\\
  $^{6}$Samsung Advanced Institute of Technology, Samsung Electronics Co., Ltd.\\
  {\small\texttt{\{ci2015.choi, quoded97, jungminko, wlaud1001, wrhee\}@snu.ac.kr}}%
}

\usepackage{colortbl}
\usepackage[most]{tcolorbox}
\usepackage{float}
\usepackage{dblfloatfix}
\usepackage[above]{placeins}
\usepackage{needspace}
\usepackage{microtype}

\definecolor{PaperBoxBackground}{gray}{0.96}
\definecolor{PaperBoxFrame}{gray}{0.45}
\newtcolorbox{paperartifactbox}[1]{
  colback=PaperBoxBackground,
  colframe=PaperBoxFrame,
  title=\textbf{#1},
  boxrule=0.5pt,
  arc=1mm,
  left=2mm,
  right=2mm,
  top=1.5mm,
  bottom=1.5mm
}

\definecolor{cvprblue}{rgb}{0.21,0.49,0.74}
\usepackage[pagebackref,breaklinks,colorlinks,allcolors=cvprblue]{hyperref}

\def\confName{CVPR}
\def\confYear{2027}

\title{VMSA: Video Memory Skill Agent for Long Video Understanding}

\begin{document}
\maketitle

\begin{abstract}
Long-video question answering requires retrieving and integrating relevant evidence from videos spanning hours or even days. Recent agentic approaches organize video information into external memories and iteratively retrieve evidence to answer questions. However, effective retrieval and reasoning remain challenging, as agents must determine what information to search for, which memory sources to access, and how to connect evidence across distant temporal intervals. To address these challenges, we propose Video Memory Skill Agent~(VMSA), a framework that jointly designs structured video memories and memory skills for long-video understanding. VMSA constructs three complementary memory components. Segment Caption Memory and Entity Event Memory serve as searchable evidence sources, while Entity-Aware Episode Memory provides broad yet detailed video context to guide evidence retrieval and reasoning. To further support effective memory utilization, VMSA induces memory skills from successful agent trajectories, incrementally refining or extending a compact skill library to guide iterative evidence retrieval, reasoning, and verification. Across seven long-video question-answering benchmarks, VMSA achieves the highest accuracy on all three week-long egocentric benchmarks, outperforming the strongest baseline by up to 9.4\% on Ego-R1 Bench, and the best or competitive performance on four additional benchmarks. Ablation studies further demonstrate the contributions of all three memory components and memory skills, with memory skills also improving temporal evidence localization.
\end{abstract}

\begin{figure*}[!t]
  \centering
  \includegraphics[width=\textwidth]{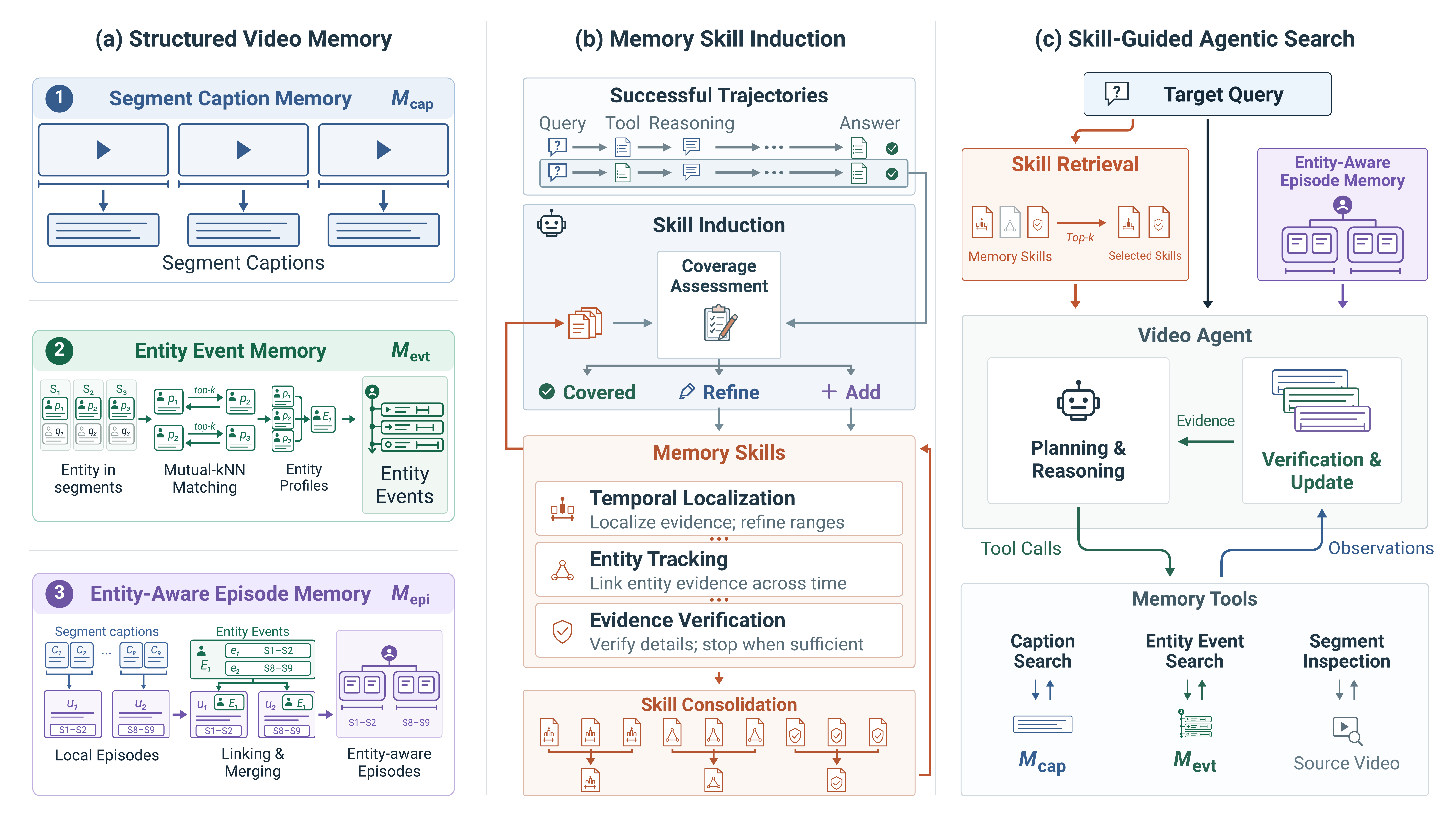}
  \caption{\textbf{Overview of VMSA.}
  \textbf{(a)} Structured video memory comprises three levels: segment captions preserving local details, entity events consolidating observations of continuing activities, and entity-aware episodes connecting related episodes across distant time points.
  \textbf{(b)} Memory skills are induced from successful source trajectories. Coverage assessment determines whether each procedure is already covered, requires skill refinement, or requires a new skill. The skill libraries are updated accordingly and periodically consolidated to reduce redundancy while preserving procedural knowledge.
  \textbf{(c)} At inference time, the video agent receives the target query, retrieved memory skills, and entity-aware episode memory as input context. The agent then gathers and verifies evidence through tool calls to generate its answer.}
  \label{fig:method_overview}
\end{figure*}
\section{Introduction}
\label{sec:introduction}
Recent advances in multimodal large language models (MLLMs) have achieved remarkable performance across a wide range of vision-language tasks~\cite{gpt5, gemini25, qwen36, qwen3vl}. However, understanding long videos remains challenging, particularly when videos span hours or even days~\cite{egolife,egor1,egomemreason}. Processing such videos directly with MLLMs incurs substantial computational overhead due to the large number of visual tokens, which may also exceed the models' available context windows~\cite{malmm,moviechat}. Moreover, long videos contain extensive information that may be irrelevant to a given question, making it difficult for MLLMs to identify fine-grained visual details and evidence dispersed across distant temporal intervals that are essential for answering the question.

To address these challenges, retrieval-augmented approaches organize information extracted from videos into searchable memories, allowing models to retrieve query-relevant evidence without processing the entire video for each question~\cite{videorag_aligned,videorag_extreme,memvid,egolife,hippomm}. These approaches use various representations, such as segment captions, hierarchical summaries, and graph-based structures, to organize video information for retrieval. More recently, agentic approaches have further developed various video memory structures and incorporated iterative reasoning and tool use to support long video question answering~\cite{avp,lenswalk,videoseek,dvd,m3agent,worldmm,egagent,videoarm,haven}. These agents iteratively retrieve relevant evidence from video memories and use the retrieved information to guide subsequent searches and reasoning. 

However, providing agents with access to extensive video memories does not necessarily ensure that they can effectively retrieve and reason over relevant evidence~\cite{yu2026longvidsearch}. In long video question answering, relevant evidence may be scattered across distant temporal intervals and multiple memory sources, making it difficult for agents to determine what information to search for, which memory sources to access, and how to connect retrieved evidence~\cite{egomemreason,scalelong}. These decisions depend not only on the question but also on the evidence gathered and reasoning performed in previous steps, further complicating iterative retrieval and reasoning~\cite{sinha2026illusion,laban2026llms}. Addressing these challenges requires more than organizing video information into searchable memories. Agents also need broad yet detailed video-wide context to help them decide what information to search for, along with strategies to guide iterative evidence retrieval, reasoning, and verification. These motivate a framework that jointly designs structured video memories and mechanisms for guiding their utilization throughout the question-answering process.

To this end, we propose Video Memory Skill Agent~(VMSA), an agentic framework that combines three types of video memories with reusable memory skills. The three memories serve distinct roles in the agent's retrieval and reasoning process. First, Segment Caption Memory stores detailed descriptions of short video segments, allowing the agent to retrieve fine-grained visual evidence and identify relevant temporal intervals. Second, Entity Event Memory consolidates observations of recurring entities into temporally grounded events, enabling retrieval of entity-specific information dispersed across different video segments. These two memories serve as searchable evidence sources accessed through search tools. Third, Entity-Aware Episode Memory connects related observations across distant temporal intervals into broader episodes to provide broad yet detailed video context. Unlike the first two memories, it is included in the agent's initial context, helping the agent determine what information to search for as its reasoning progresses.

To further guide evidence retrieval and reasoning, VMSA introduces memory skills that capture effective strategies for selecting memory sources, connecting retrieved observations, verifying key observations, and determining when sufficient evidence has been collected. To construct the compact skill library, it sequentially analyze successful agent trajectories and assess whether their procedural knowledge is already covered by existing skills, refining existing skills or adding new ones as needed to capture strategies while minimizing redundancy. At inference time, the agent receives the target question, Entity-Aware Episode Memory, and relevant memory skills as its initial context. While Entity-Aware Episode Memory provides video-wide context to guide evidence searches, memory skills provide transferable strategies for retrieving and reasoning over evidence. Guided by both, the agent iteratively searches Segment Caption Memory and Entity Event Memory and, when necessary, directly inspects relevant segments of the source video to gather and verify evidence before generating an answer.

We evaluate VMSA on a comprehensive suite of long-video question-answering benchmarks, including three egocentric benchmarks with week-long videos and four benchmarks with videos ranging from tens of minutes to over an hour. Compared with a broad range of baselines, including proprietary MLLMs, specialized long video models, retrieval-augmented methods, and recent agentic frameworks, VMSA achieves the best performance on all three week-long egocentric benchmarks, outperforming the strongest baselines by up to 9.4\% on Ego-R1 Bench~\cite{egor1}. It also achieves the best or competitive performance on the remaining four benchmarks. Ablation studies show that all three memory components and memory skills contribute to answer accuracy, while memory skills also improve temporal evidence localization. These results highlight the importance of jointly designing structured video memories and effective mechanisms for guiding their utilization.

\section{Related Work}
\label{sec:related-work}

\subsection{Long-Video Understanding}
\label{subsec:related_work_long_video}

Recent advances in MLLMs have enabled video understanding through sampled video frames~\cite{gpt5,gemini25,qwen36}. To improve long-video understanding, specialized models have been developed to process extended video sequences and enhance temporal reasoning. Some approaches employ efficient visual token processing and compression~\cite{videochatflash}, while others use post-training techniques, including reinforcement learning, to improve multi-step reasoning and temporal grounding~\cite{longvilar1,videorft,videor1,deepvideor1,rewatchr1,timer1,videoitg,videorts}. However, processing long videos directly remains challenging due to the large number of visual tokens and the difficulty of capturing fine-grained evidence across distant temporal intervals.

Retrieval-augmented methods address these challenges by organizing video information into searchable representations, such as captions, hierarchical summaries, and knowledge graphs~\cite{videorag_aligned,videorag_extreme,memvid,egolife,hippomm}. These methods allow models to retrieve relevant evidence without repeatedly processing entire videos. More recently, agentic frameworks have incorporated iterative reasoning and tool use to progressively gather and examine video evidence~\cite{avp,lenswalk,videoseek,dvd,m3agent,worldmm,egagent,videoarm,haven}. These approaches enable agents to adapt their searches based on intermediate observations and reasoning, rather than relying solely on fixed retrieval procedures.

\subsection{Agentic Video Frameworks}
\label{subsec:related_work_video_agents}
Agentic approaches to long-video understanding iteratively gather evidence and use intermediate reasoning to guide subsequent searches. Early methods employ LLMs to select query-relevant frames or retrieve information from temporal and object memories~\cite{videoagent_llm,videoagent_memory}. Hierarchical approaches, such as VideoTree and VideoLucy, explore relevant video segments through tree expansion or memory-based backtracking~\cite{videotree,videolucy}. More recent frameworks, including Active Video Perception (AVP), LensWalk, VideoSeek, and Deep Video Discovery (DVD), dynamically adjust search locations, observation scope, or temporal resolution based on intermediate reasoning~\cite{avp,lenswalk,videoseek,dvd}.

Recent studies have also developed structured video memories to facilitate agentic retrieval and reasoning. M3-Agent employs entity-centric memory~\cite{m3agent}, WorldMM uses multi-scale episodic, semantic, and visual memories~\cite{worldmm}, and EGAgent constructs entity graphs to connect evidence across temporal intervals~\cite{egagent}. VideoARM and HAVEN further integrate hierarchical memories with audio transcripts for iterative retrieval~\cite{videoarm,haven}. However, organizing video information into structured memories alone does not ensure effective agentic retrieval and reasoning, as agents must determine what information to search for and how to connect evidence across multiple memory sources. Motivated by this challenge, VMSA jointly designs complementary video memories and a memory-skill construction framework to guide evidence retrieval, reasoning, and verification.

\section{Method}
\label{sec:method}

Our method, VMSA, is a long-video question-answering agent that combines structured video memory with memory tools and reusable skills to effectively gather evidence from videos spanning hours to days. We first describe the construction of structured video memory in Section~\ref{sec:structured-memory}, followed by the memory tools and agentic search process in Section~\ref{sec:agentic-search}. We then describe the skill induction process, which extracts reusable procedural knowledge from successful trajectories, in Section~\ref{sec:memory-skills}. Finally, we explain how VMSA iteratively gathers evidence and generates answers at inference time in Section~\ref{sec:skill-guided-search}. Figure~\ref{fig:method_overview} provides an overview of our method.

\subsection{Structured Video Memory}
\label{sec:structured-memory}
Our structured video memory consists of three levels: Segment Caption Memory, Entity Event Memory, and Entity-Aware Episode Memory. In this section, we describe each memory component and its construction process, with further implementation details provided in Appendix~\ref{sec:appendix-memory-construction}. 

\subsubsection{Segment Caption Memory}
\label{sec:segment-caption-memory}
Segment Caption Memory consists of detailed captions of video segments and their corresponding temporal ranges. To construct this memory, we first divide a long video into short temporal segments. For each segment $s_i$, we uniformly sample frames and provide them to a memory-construction vision-language model (VLM) to generate a caption $C_i$. Each caption describes the visual content of the segment, including participants, objects, actions, interactions, and state changes. The resulting memory is defined as
\begin{equation}
    \mathcal{M}_{\mathrm{cap}}
    = \{(C_i,\tau_i)\}_{i=1}^{N_s},
    \label{eq:segment-caption-memory}
\end{equation}
where $\tau_i$ denotes the temporal range of segment $s_i$, and $N_s$ is the total number of segments.

\subsubsection{Entity Event Memory}
\label{sec:entity-event-memory}
Entity Event Memory consists of entities and their associated events, each describing an activity, interaction, or state change involving an entity. To construct this memory, we progressively merge entity profiles within and across sliding temporal windows, then consolidate their associated events.

Specifically, for each video segment $s_i$, the memory-construction VLM extracts local entity profiles from uniformly sampled frames, where each profile records observable attributes, actions, relations, and state changes. Within each sliding temporal window, we construct a mutual-$k$NN graph~\cite{von2007tutorial} over profile embeddings to identify candidate groups of profiles referring to the same entity. The VLM verifies these groups and merges profiles referring to the same entity by aggregating their attributes. This merging process is periodically repeated across profiles accumulated from previous windows, followed by a final video-level merging pass, yielding $N_e$ entities.

For each resulting entity $E_k$, we consolidate descriptions of the same continuing event by removing duplicates and extending its temporal range to cover all supporting observations. The resulting Entity Event Memory is defined as
\begin{equation}
    \mathcal{M}_{\mathrm{evt}}
    = \bigcup_{k=1}^{N_e}\{(E_k,e_{k,1}, e_{k,2}, \dots, e_{k,J_k})\},
    \label{eq:entity-event-memory}
\end{equation}
where $N_e$ is the number of entities, $J_k$ is the number of events associated with $E_k$, and $e_{k,j}$ denotes the $j$-th event of $E_k$, comprising an event description and its temporal range.

\subsubsection{Entity-Aware Episode Memory}
\label{sec:entity-aware-episodes}

Entity-Aware Episode Memory consists of a global video summary and entity-aware episodes that connect related observations of the same entity across temporally dispersed intervals. To construct this memory, we first merge temporally adjacent Segment Captions describing the same ongoing activity or interaction into local episodes. The memory-construction VLM then uses Entity Events to identify local episodes involving the same entity and merges semantically related episodes across different temporal intervals into entity-aware episodes. Finally, the VLM generates a global video summary using all the entity-aware episodes. 
The resulting Entity-Aware Episode Memory is defined as
\begin{equation}
	\mathcal{M}_{\mathrm{epi}} = \{G\} \cup \bigcup_{k=1}^{N_e} \{u_{k,j}\}_{j=1}^{L_k},
\end{equation}
where $G$ denotes the global video summary, $N_e$ is the number of entities, $L_k$ is the number of entity-aware episodes associated with entity $E_k$, and $u_{k,j}$ denotes the $j$-th entity-aware episode of $E_k$, comprising an episode description and its supporting temporal ranges. Unlike Entity Events, which describe entity-specific activities, interactions, and state changes, Entity-Aware Episodes provide broader contextual descriptions by integrating semantically related local episodes across different temporal intervals. Entity-Aware Episode Memory thus provides a broad yet detailed overview of the video that helps the agent determine what information to search for as its reasoning progresses.

\subsection{Search and Verification Tools}
\label{sec:agentic-search}

Our video agent iteratively gathers and examines evidence using tools that operate over structured video memory and the source video. While Entity-Aware Episode Memory is included in the agent's initial context, Segment Caption Memory and Entity Event Memory are accessed through search tools, and the source video is directly inspected through a verification tool when further verification is needed. In this section, we describe these three tools, with retrieval equations and additional details provided in Appendix~\ref{sec:appendix-memory-search}.

\vspace{4pt}
\noindent\textbf{Caption Search.} 
Caption Search retrieves segment captions and their temporal ranges from Segment Caption Memory. The agent formulates a search query based on the question and accumulated evidence. Using embedding-based retrieval, the tool identifies query-relevant segment captions and returns them with their corresponding temporal ranges. The retrieved captions provide fine-grained visual descriptions, while their temporal ranges indicate candidate intervals for further inspection.

\vspace{4pt}
\noindent\textbf{Entity Event Search.}
Entity Event Search retrieves entity-related evidence from Entity Event Memory. The agent formulates a query specifying a target entity and relevant event information. Using embedding-based retrieval, the tool identifies relevant events and returns their descriptions, supporting temporal ranges, and associated entity profiles. This allows the agent to retrieve observations of the same entity across temporally distant intervals.

\vspace{4pt}
\noindent\textbf{Segment Inspection.}
Segment Inspection examines visual details directly from the source video when evidence retrieved from structured video memory is insufficient. The agent formulates a verification query specifying the information to examine and selects relevant temporal intervals for inspection. The tool inspects video frames within these intervals and returns observations that help resolve uncertainties or verify candidate interpretations.

\subsection{Memory Skills}
\label{sec:memory-skills}
We induce memory skills from successful agent trajectories to guide the agent's iterative reasoning and tool use across multiple memory sources and the source video. Each skill provides structured guidance on when to apply it, how to retrieve and verify evidence, what information to retain, how to reason over accumulated evidence, when to terminate the search, and what common mistakes to avoid. To construct a skill library, an induction LLM sequentially processes successful trajectories, assessing whether the procedural knowledge from each trajectory is already covered by existing skills and refining existing skills or adding new ones as needed. The memory-skill template and an example of an induced skill are provided in Figure~\ref{fig:appendix_skill_example} of Appendix~\ref{sec:appendix-skill-induction}, with further details provided in Appendix~\ref{sec:appendix-skill-induction}.

\vspace{6pt}
\noindent\textbf{Collecting Successful Trajectories.}
To encourage generalization, we collect successful trajectories from source video benchmarks separate from those used for evaluation. We run the agent on these benchmarks using structured video memories and the search and verification tools, retaining trajectories whose final predictions are correct. For each successful trajectory, we preserve the agent's reasoning and tool calls while omitting raw tool results to reduce context length. 

\vspace{6pt}
\noindent\textbf{Incremental Skill Library Update.}
Starting from an empty skill library $\mathcal{P}_0=\varnothing$, we process successful trajectories sequentially. At step $i$, the induction LLM receives the $i$-th successful trajectory $\xi_i$ and the current library $\mathcal{P}_{i-1}$, and updates the library as follows:

\begin{equation}
    \mathcal{P}_i =
    \operatorname{Induce}
    \left(\xi_i, \,\mathcal{P}_{i-1}\right),
    \label{eq:skill-library-update}
\end{equation}
where $\operatorname{Induce}(\cdot)$ denotes the library-update operation performed by the induction LLM. The LLM assesses whether existing skills cover the retrieval, reasoning, and verification procedures demonstrated in the trajectory. If they do, the library remains unchanged. Otherwise, the LLM refines an existing skill or adds a new one, depending on whether the missing procedural knowledge can be incorporated into an existing skill. Throughout this process, trajectory-specific information is abstracted away to extract transferable procedures that generalize across queries and videos.

\vspace{6pt}
\noindent\textbf{Periodic Skill Library Consolidation.}
After every 100 processed trajectories, the induction LLM is prompted to reduce redundancy in the library by merging skills that capture the same underlying procedure. The resulting skill library $\mathcal{P}$ is used during evaluation on the target benchmarks.

\subsection{Skill-Guided Agentic Search}
\label{sec:skill-guided-search}
At inference time, the agent retrieves the top-$K_s$ query-relevant memory skills from the induced skill library $\mathcal{P}$ using embedding-based similarity. The agent receives the target question, retrieved memory skills, and the entire Entity-Aware Episode Memory as its initial context. Guided by this context, the agent iteratively reasons about the question and invokes Caption Search, Entity Event Search, or Segment Inspection as needed to gather evidence and verify candidate claims. At each iteration, the agent incorporates the tool observations into its accumulated context and determines whether further evidence is needed. This process continues until the agent decides that sufficient evidence has been collected to generate a final answer or reaches the maximum number of iterations. Figure~\ref{fig:appendix-vmsa-example} in the Appendix provides a qualitative example of VMSA's inference process, including query-relevant memory skill retrieval, iterative evidence gathering and reasoning through search and verification tools, and final answer generation.

\section{Experiments}
\label{sec:experiments}

\subsection{Experimental Setup}
%
\noindent\textbf{Benchmarks.}
We evaluate VMSA on seven multiple-choice long-video question-answering benchmarks covering diverse video domains, durations, and question types.  The three week-long egocentric benchmarks are EgoLifeQA~\cite{egolife}, Ego-R1 Bench~\cite{egor1}, and EgoMemReason~\cite{egomemreason}, which contain egocentric videos recorded over seven days, with an average of 44.3 hours of footage per participant. The other four benchmarks are Video-MME~(long split)~\cite{videomme}, LongVideoBench~(long split)~\cite{longvideobench}, LVBench~\cite{lvbench}, and HippoVlog~\cite{hippomm}, with average video durations ranging from approximately 24 to 67 minutes. Together, these benchmarks assess VMSA's ability to identify fine-grained visual details, locate relevant events, and retrieve and integrate evidence across different temporal intervals to answer questions. The week-long egocentric benchmarks provide a particularly challenging setting due to their substantially longer videos. Benchmark statistics, including the number of source videos or participant streams, the number of questions, and average video durations, are provided in Appendix~\ref{sec:dataset}.
\paragraph{Baselines.}
We compare VMSA with four categories of baselines. \emph{Base models} include Gemini-2.5-Pro and Gemini-2.5-Flash~\cite{gemini25}, Gemini-3.1-Pro~\cite{gemini31pro}, o3~\cite{openai_o3}, GPT-4o~\cite{gpt4o}, GPT-5~\cite{gpt5}, Qwen3-VL-8B~\cite{qwen3vl}, and Qwen3.6-27B~\cite{qwen36}. \emph{Long-video LLMs} include VideoChat-Flash~\cite{videochatflash}, Time-R1~\cite{timer1}, Video-RTS~\cite{videorts}, LongVA~\cite{longva}, StreamingVLM~\cite{streamingvlm}, InternVideo-2.5~\cite{internvideo25}, VideoLLaMA3~\cite{videollama3}, and Molmo2~\cite{molmo2}. \emph{Retrieval-augmented methods} include LightRAG~\cite{lightrag}, HippoRAG~\cite{hipporag}, Video-RAG~\cite{videorag_aligned}, MemVid~\cite{memvid}, EgoRAG~\cite{egolife}, and HippoMM~\cite{hippomm}. \emph{Agentic video frameworks} include Logic-in-Frames~\cite{logicinframes}, MR.\ Video~\cite{mrvideo}, VideoTree~\cite{videotree}, VideoLucy~\cite{videolucy}, AVP~\cite{avp}, LensWalk~\cite{lenswalk}, VideoSeek~\cite{videoseek}, Ego-R1~\cite{egor1}, M3-Agent~\cite{m3agent}, EGAgent~\cite{egagent}, WorldMM~\cite{worldmm}, SiLVR~\cite{silvr}, and DVD~\cite{dvd}. Unless otherwise specified, results for VideoChat-Flash, Time-R1, Video-RTS, LightRAG, HippoRAG, Video-RAG, EgoRAG, HippoMM, Ego-R1, and M3-Agent are taken from WorldMM~\cite{worldmm}, while results for the other baselines are taken from their original papers.

\paragraph{Implementation Details.}
For structured video memory construction, we use Qwen3.5-9B~\cite{qwen35} as the memory-construction VLM, dividing videos into 60-second segments for the three week-long egocentric benchmarks and 16-second segments for the remaining four benchmarks. To reduce computational cost for the week-long egocentric benchmarks, we reuse the segment captions released by WorldMM~\cite{worldmm} instead of generating new ones. For memory skill induction, we use GPT-5.6~\cite{gpt56} as the induction LLM and collect successful agent trajectories without memory-skill guidance. To evaluate the transferability of the induced skills across benchmarks, we construct separate skill libraries using trajectories from benchmarks outside each target evaluation group. Specifically, the library for the week-long egocentric benchmarks is constructed from ScaleLong~\cite{scalelong}, VideoOdyssey-V~\cite{videoodyssey}, Video-MME, LongVideoBench, LVBench, and EgoSchema~\cite{egoschema}, while the library for the remaining four benchmarks is constructed from ScaleLong and VideoOdyssey-V. For skill-guided agentic search, we use Qwen3.6-27B~\cite{qwen36} as the agent model, retrieve the top-$K_s=3$ query-relevant memory skills, and allow up to 10 iterations per question. To reduce inference cost, we disable Segment Inspection for the week-long egocentric benchmarks. Further details are provided in Appendix~\ref{sec:appendix-implementation}.

\begin{table}[ht]
  \centering
  \caption{
    Comparison on EgoLifeQA~\cite{egolife} and Ego-R1 Bench~\cite{egor1}.
    We report multiple-choice accuracy~(\%).
    Best and second-best results are shown in \textbf{bold} and \underline{underlined}, respectively.
    }
  \label{tab:egolifeqa_egor1}
  \small
  \setlength{\tabcolsep}{4pt}
  \resizebox{\columnwidth}{!}{%
  \begin{tabular}{@{}lcc@{}}
    \toprule
    Method & EgoLifeQA & Ego-R1 Bench \\
    \midrule
    \multicolumn{3}{l}{\textit{Base Models}} \\
    Gemini-2.5-Pro~\cite{gemini25}   & 46.4 & 46.7 \\
    GPT-5~\cite{gpt5}                & 48.6 & 46.3 \\
    Qwen3.6-27B~\cite{qwen36}        & 34.0 & 42.3 \\
    \midrule
    \multicolumn{3}{l}{\textit{Long-Video LLMs}} \\
    VideoChat-Flash~\cite{videochatflash} & 34.2 & 42.7 \\
    Time-R1~\cite{timer1}            & 48.8 & 48.0 \\
    Video-RTS~\cite{videorts}        & 48.2 & 48.0 \\
    \midrule
    \multicolumn{3}{l}{\textit{Retrieval-Augmented Methods}} \\
    LightRAG~\cite{lightrag}         & 48.8 & 52.3 \\
    HippoRAG~\cite{hipporag}         & 59.6 & 56.0 \\
    Video-RAG~\cite{videorag_aligned} & 55.4 & 49.7 \\
    EgoRAG~\cite{egolife}            & 52.0 & 49.0 \\
    HippoMM~\cite{hippomm}           & 54.6 & 53.0 \\
    \midrule
    \multicolumn{3}{l}{\textit{Agentic Video Frameworks}} \\
    Ego-R1~\cite{egor1} (3B)         & 53.0 & 52.0 \\
    M3-Agent~\cite{m3agent} (7B)     & 53.5 & 52.0 \\
    EGAgent~\cite{egagent} (Gemini-2.5-Pro) & 57.5 & -  \\
    WorldMM~\cite{worldmm} (GPT-5)   & \underline{65.6} & \underline{65.3} \\
    \midrule
    \textbf{VMSA (Qwen3.6-27B)} & \textbf{66.8} & \textbf{74.7} \\
    \bottomrule
  \end{tabular}%
  }
  \par\vspace{4pt}
\end{table}
\begin{table}[!htbp]
  \centering
  \caption{
  Comparison on the EgoMemReason benchmark~\cite{egomemreason}. We report multiple-choice accuracy (\%). Best and second-best results are shown in \textbf{bold} and \underline{underlined}, respectively.}
  \label{tab:egomemreason}
  \normalsize
  \setlength{\tabcolsep}{4pt}
  \begin{tabular}{@{}lc@{}}
    \toprule
    Method & Overall \\
    \midrule
    \multicolumn{2}{l}{\textit{Base Models}} \\
    Qwen3-VL-8B~\cite{qwen3vl}        & 29.6 \\
    GPT-5~\cite{gpt5}                 & 27.8 \\
    Gemini-3.1-Pro~\cite{gemini31pro} & \underline{37.4} \\
    Qwen3.6-27B~\cite{qwen36}         & 30.6 \\
    \midrule
    \multicolumn{2}{l}{\textit{Long-Video LLMs}} \\
    LongVA-7B~\cite{longva}           & 20.6 \\
    StreamingVLM~\cite{streamingvlm}  & 24.2 \\
    InternVideo-2.5-8B~\cite{internvideo25} & 25.6 \\
    VideoLLaMA3-8B~\cite{videollama3} & 30.0 \\
    Molmo2-8B~\cite{molmo2}           & 33.2 \\
    \midrule
    \multicolumn{2}{l}{\textit{Agentic Video Frameworks}} \\
    SiLVR~\cite{silvr} (GPT-5)        & 22.4 \\
    Ego-R1~\cite{egor1} (Qwen2.5-3B)  & 25.8 \\
    WorldMM~\cite{worldmm} (GPT-5)    & 30.6 \\
    AVP~\cite{avp} (Gemini-2.5-Flash) & 34.0 \\
    \midrule
    \textbf{VMSA (Qwen3.6-27B)} & \textbf{38.6} \\
    \bottomrule
  \end{tabular}
  \par\vspace{4pt}
\end{table}

\begin{table*}[!t]
  \centering
  \caption{Comparison on Video-MME (L)~\cite{videomme}, LongVideoBench
           (L)~\cite{longvideobench}, LVBench~\cite{lvbench}, and
           HippoVlog~\cite{hippomm}. We report accuracy~(\%) on
           multiple-choice questions. The best and second-best
           results are shown in \textbf{bold} and \underline{underlined},
           respectively. }
  \label{tab:longvideo_benchmarks}

  \small
  \setlength{\tabcolsep}{4pt}
  \begin{tabular*}{\linewidth}{@{\extracolsep{\fill}}lcccc@{}}
    \toprule
    Method & Video-MME (L) & LongVideoBench (L) & LVBench & HippoVlog \\
    \midrule
    \multicolumn{5}{l}{\textit{Base Models}} \\
    Gemini-2.5-Flash~\cite{gemini25}    & 69.1 & 61.8 & 56.7 & -  \\
    o3~\cite{openai_o3}                 & 64.7 & 60.6 & 57.1 & -  \\
    GPT-4o~\cite{gpt4o}                 & 65.3 & 60.9 & 48.9 & -  \\
    GPT-5~\cite{gpt5}                   & 74.3 & -  & 60.4 & 75.7 \\
    Qwen3.6-27B~\cite{qwen36}           & 65.0 & 62.2 & 53.5 & 78.1 \\
    \midrule
    \multicolumn{5}{l}{\textit{Retrieval-Augmented Methods}} \\
    MemVid~\cite{memvid}                & 55.0 & -  & 44.4 & -  \\
    EgoRAG~\cite{egolife}              & 41.1 & -  & 32.2 & 57.5 \\
    HippoMM~\cite{hippomm}             & 41.6 & -  & 38.2 & 71.9 \\
    \midrule
    \multicolumn{5}{l}{\textit{Agentic Video Frameworks}} \\
    Ego-R1~\cite{egor1} (Qwen2.5-3B)   & 42.7 & -  & 34.1 & 58.8 \\
    M3-Agent~\cite{m3agent} (7B)       & 55.3 & -  & 49.3 & 65.5 \\
    Logic-in-Frames~\cite{logicinframes} (GPT-4o) & 55.2 & 54.2 & -  & -  \\
    MR.\ Video~\cite{mrvideo} (GPT-4o) & 61.8 & 61.6 & 60.8  & -  \\
    VideoTree~\cite{videotree} (GPT-4)  & 54.2 & -  & -  & -  \\
    VideoLucy~\cite{videolucy} (DeepSeek-R1) & 66.8 & -  & 58.8 & -  \\
    AVP~\cite{avp} (Gemini-2.5-Flash)   & \textbf{76.7} & 65.5 & 63.8 & -  \\
    LensWalk~\cite{lenswalk} (GPT-5)    & 66.9 & 68.8 & 69.2 & -  \\
    VideoSeek~\cite{videoseek} (GPT-5)  & 70.1 & \underline{73.5} & 68.4 & -  \\
    WorldMM~\cite{worldmm} (GPT-5)      & \underline{76.6} & -  & 61.9 & \underline{78.3} \\
    DVD~\cite{dvd} (o3)                 & 67.3 & 68.6 & \textbf{74.2} & -  \\
    \midrule
    \textbf{VMSA (Qwen3.6-27B)} & 76.1 & \textbf{74.1} & \underline{72.8} & \textbf{82.1} \\
    \bottomrule
  \end{tabular*}
\end{table*}

\subsection{Main Results}

\subsubsection{Week-Long Egocentric Benchmarks}
\label{sec:main_egolife}

\paragraph{EgoLifeQA and Ego-R1 Bench.}
Table~\ref{tab:egolifeqa_egor1} presents the results on EgoLifeQA and Ego-R1 Bench. VMSA achieves the highest accuracy on both benchmarks, outperforming all compared baselines across the four categories. Specifically, VMSA achieves 66.8\% accuracy on EgoLifeQA and 74.7\% on Ego-R1 Bench, surpassing the second-best method, WorldMM, by 1.2\% and 9.4\%, respectively. These results demonstrate the effectiveness of VMSA in retrieving and reasoning over relevant evidence from week-long egocentric videos. Detailed results for each question category defined in the benchmarks are provided in Table~\ref{tab:appendix_egolife_category} of Appendix~\ref{sec:Category-Wise Results on Week-Long Egocentric Benchmarks}.

\paragraph{EgoMemReason.}
Compared with the other week-long egocentric benchmarks, EgoMemReason focuses more on questions that require retrieving and integrating evidence from multiple video segments across distant temporal intervals. On average, each question requires 5.1 distinct evidence segments~\cite{egomemreason}. As shown in Table~\ref{tab:egomemreason}, VMSA achieves the highest overall accuracy of 38.6\%, outperforming the second-best baseline, Gemini-3.1-Pro, by 1.2\%. Among agentic video frameworks, VMSA outperforms the strongest baseline, AVP, by 4.6\%. Detailed results for each reasoning type defined in EgoMemReason are provided in Table~\ref{tab:appendix_egomemreason_category} of Appendix~\ref{sec:Reasoning-Type Analysis on EgoMemReason}.

\subsubsection{Other Long-Video Benchmarks}
\label{sec:main_longvideo}
Table~\ref{tab:longvideo_benchmarks} presents the results on Video-MME (L), LongVideoBench (L), LVBench, and HippoVlog, with average video durations ranging from approximately 24 to 67 minutes. VMSA achieves the best or competitive performance across all four benchmarks. Specifically, VMSA achieves the highest accuracy on LongVideoBench and HippoVlog, outperforming the second-best methods by 0.6\% and 3.8\%, respectively. On Video-MME and LVBench, VMSA remains competitive with the strongest baselines, achieving accuracies of 76.1\% and 72.8\%, respectively. Overall, these results demonstrate the effectiveness of VMSA across diverse long-video understanding benchmarks, complementing its strong performance on the three week-long egocentric benchmarks.

\section{Analysis}
\label{sec:analysis}

\subsection{Effect of Memory Components}
\label{sec:abl_retrieval_views}
We evaluate the contribution of each memory component by individually removing it from VMSA. As shown in Table~\ref{tab:ablation_memory_components}, using all three memories achieves the highest accuracy across LongVideoBench (L), HippoVlog, and EgoLifeQA. Removing Segment Caption Memory leads to the largest performance drops across all three benchmarks, ranging from 5.5\% to 9.2\%, while removing Entity Event Memory also consistently reduces accuracy. These results highlight the importance of complementary searchable evidence sources that provide fine-grained descriptions of individual video segments and entity-specific information dispersed across temporal intervals. Removing Entity-Aware Episode Memory also decreases accuracy across all three benchmarks, supporting the benefit of providing broad yet detailed video context in the agent's initial context to guide evidence retrieval and reasoning. Overall, these results demonstrate that the three memory components serve complementary roles and jointly contribute to effective long-video understanding.

\begin{table}[!tbp]
  \centering
  \caption{Ablation of memory components in VMSA on LongVideoBench (L), HippoVlog, and EgoLifeQA. Caption, Event, Episode refer to Segment Caption Memory, Entity Event Memory, and Entity-Aware Episode Memory, respectively.}
  \label{tab:ablation_memory_components}
  \setlength{\tabcolsep}{6pt}
  \resizebox{\columnwidth}{!}{%
  \begin{tabular}{ccc ccc}
    \toprule
    Caption & Event & Episode & LongVideoBench (L) & HippoVlog & EgoLifeQA \\
    \midrule
    -          & \checkmark & \checkmark & 68.6          & 73.9          & 57.6          \\
    \checkmark & -          & \checkmark & 72.5          & 74.3          & 63.6          \\
    \checkmark & \checkmark & -          & 70.6          & 75.5          & 62.4          \\
    \checkmark & \checkmark & \checkmark & \textbf{74.1} & \textbf{82.1} & \textbf{66.8} \\
    \bottomrule
  \end{tabular}%
  }
\end{table}

\begin{table}[!htbp]
  \centering
  \caption{Ablation of memory skills. We report average best-turn tIoG and tIoU, and answer accuracy~(\%). For ScaleLong and VideoOdyssey-V, memory skills are induced from Video-MME~(L), LongVideoBench~(L), LVBench, and HippoVlog. }
  \label{tab:analysis_skill_grounding}
  \setlength{\tabcolsep}{3pt}
  \resizebox{\columnwidth}{!}{%
  \begin{tabular}{l c ccc}
    \toprule
    Dataset & Memory Skills & Avg. tIoG & Avg. tIoU & Accuracy \\
    \midrule
    LVBench
      & -          & 0.447          & 0.170           & 71.8          \\
    LVBench                              
      & \checkmark & \textbf{0.522} & \textbf{0.184}  & \textbf{72.8} \\
    \midrule                             
    Ego-R1 Bench                         
      & -          & 0.597          & 0.044           & 70.7          \\
    Ego-R1 Bench                         
      & \checkmark & \textbf{0.641} & \textbf{0.048}  & \textbf{74.7} \\
    \midrule                             
    ScaleLong                            
      & -          & 0.441          & 0.127           & 67.7          \\
    ScaleLong                            
      & \checkmark & \textbf{0.466} & \textbf{0.132}  & \textbf{69.1} \\
    \midrule                             
    VideoOdyssey-V                       
      & -          & 0.569          & 0.265           & 65.5          \\
    VideoOdyssey-V                       
      & \checkmark & \textbf{0.582} & \textbf{0.275}  & \textbf{66.4} \\
    \bottomrule
  \end{tabular}%
  }
\end{table}

\subsection{Effect of Memory Skills}
\label{sec:ana_memory_skills}
We evaluate the effect of memory skills on temporal evidence localization and answer accuracy by comparing VMSA with and without memory skills. For benchmarks providing ground-truth evidence intervals, we compute temporal Intersection over Ground Truth~(tIoG) and Intersection over Union~(tIoU) between the temporal intervals observed by the agent at each iteration and the annotated evidence intervals. We calculate average best-turn tIoG and tIoU by selecting the highest value across iterations for each question and then averaging across questions.

As shown in Table~\ref{tab:analysis_skill_grounding}, memory skills consistently improve tIoG across all four benchmarks, while also improving tIoU to a lesser extent. These results suggest that memory skills help the agent cover a greater proportion of ground-truth evidence without merely expanding the observed temporal intervals to include irrelevant content. These improvements in temporal evidence localization may help explain the higher answer accuracy achieved with memory skills across all four benchmarks. Additional answer accuracy results for benchmarks without ground-truth evidence intervals are provided in Table~\ref{tab:appendix_ablation_memory_skills} of the Appendix.

\begin{table}[!htbp]
  \centering
  \caption{Comparison of skill induction strategies. We report skill library size, measured by the number of induced skills or experience entries, and answer accuracy~(\%) on LongVideoBench~(L) and LVBench.}
  \label{tab:analysis_generic_skills}
  \small
  \setlength{\tabcolsep}{4pt}
  \resizebox{\columnwidth}{!}{%
  \begin{tabular}{lccc}
    \toprule
    Skill Induction Strategy & Skill Size & LongVideoBench (L) & LVBench \\
    \midrule
    No Memory Skills                              & 0     & 71.3          & 71.8 \\
    AWM~\cite{awm}                                & 266   & 63.1          & 68.2 \\
    MemP~\cite{memp}                              & 2,181 & 71.6          & 70.7 \\
    ReasoningBank~\cite{ouyang2026reasoningbank}  & 3,325 & 71.8          & 71.2 \\
    Trace2Skill~\cite{trace2skill}                & 1     & 69.7          & 72.1 \\
    VMSA (Ours)                                    & 15    & \textbf{74.1} & \textbf{72.8} \\
    \bottomrule
  \end{tabular}%
  }
\end{table}

\subsection{Effect of Skill Induction Strategies}
\label{sec:ana_generic_skills}
We compare our skill induction approach with four existing methods: AWM~\cite{awm}, MemP~\cite{memp}, ReasoningBank~\cite{ouyang2026reasoningbank}, and Trace2Skill~\cite{trace2skill}. We evaluate each method by replacing VMSA's skill induction approach with the alternative while keeping all other components unchanged. As shown in Table~\ref{tab:analysis_generic_skills}, VMSA achieves the highest accuracy on both LongVideoBench (L) and LVBench.

The methods also differ substantially in the sizes of their induced skill libraries. AWM, MemP, and ReasoningBank produce hundreds to thousands of skills or experience entries, while Trace2Skill consolidates trajectory-derived knowledge into a single skill. In contrast, VMSA constructs a compact library of 15 skills by comparing each new trajectory with the existing library, selectively adding or refining skills to incorporate new procedural knowledge, and periodically consolidating redundant skills. These results suggest that VMSA's incremental skill induction approach may help maintain a compact library of transferable skills, potentially contributing to its strong performance on both benchmarks.

\subsection{Analysis of Search-Turn Budget}
\label{sec:abl_turn_budget}
We examine how the maximum search-turn budget affects VMSA's answer accuracy. Table~\ref{tab:ablation_turn_budget} shows that increasing the budget from 3 to 7 turns improves accuracy on all three benchmarks. The gains saturate at 7 turns on LVBench and Video-MME~(L). On LongVideoBench~(L), accuracy improves at every budget increase, rising from 62.8\% at 3 turns to 71.1\% at 7 turns and 74.1\% at 10 turns. VMSA uses an average of 4.60 turns per question on LongVideoBench~(L), the highest among the seven benchmarks, suggesting that some questions benefit from additional turns to gather sufficient evidence.



\begin{table}[!tbp]
  \centering
  \caption{Mean and sample standard deviation of search turns per question under the default 10-turn budget.}
  \label{tab:analysis_search_turns}
  \footnotesize
  \setlength{\tabcolsep}{6pt}
  \begin{tabular*}{\columnwidth}{@{\extracolsep{\fill}}lrr@{}}
    \toprule
    Dataset & Avg. Turn & Std \\
    \midrule
    Video-MME (L)     & 3.90 & 2.56 \\
    LongVideoBench (L) & 4.60 & 2.94 \\
    LVBench           & 3.23 & 2.33 \\
    HippoVlog         & 2.60 & 1.86 \\
    EgoLifeQA         & 2.42 & 1.88 \\
    EgoMemReason      & 2.07 & 2.18 \\
    Ego-R1 Bench      & 2.35 & 1.76 \\
    \bottomrule
  \end{tabular*}
\end{table}

\subsection{Effect of Agent Models}
\label{sec}
We evaluate the effect of replacing VMSA's default agent model with alternative models while keeping all other components unchanged. As shown in Table~\ref{tab:analysis_model_scaling}, GPT-5 achieves the highest accuracy across all four benchmarks. Notably, replacing the default Qwen3.6-27B with GPT-5 improves accuracy on Video-MME~(L) from 76.1\% to 77.4\%, surpassing all baselines in Table~\ref{tab:longvideo_benchmarks}. These results suggest that VMSA can benefit from using stronger agent models while remaining effective across different models.



\begin{table}[!tbp]
  \centering
  \caption{Effect of agent models on answer accuracy~(\%). For the GPT-5 configuration on Video-MME~(L), we use Qwen3.6-27B for Segment Inspection due to the high cost of using GPT-5.}
  \label{tab:analysis_model_scaling}
  \setlength{\tabcolsep}{4pt}
  \resizebox{\columnwidth}{!}{%
  \begin{tabular}{l cccc}
    \toprule
    Agent Model & Video-MME (L) & EgoLifeQA & EgoMemReason & Ego-R1 Bench \\
    \midrule
    Qwen3.5-9B          & 66.9          & 55.8          & 35.4          & 65.7          \\
    Qwen3.5-122B-A10B   & 73.4          & 62.8          & 38.4          & 71.3          \\
    Qwen3.6-27B~(Default)         & 76.1          & 66.8          & 38.6          & 74.7          \\
    GPT-5               & \textbf{77.4} & \textbf{70.0} & \textbf{41.4} & \textbf{75.0} \\
    \bottomrule
  \end{tabular}%
  }
\end{table}

\section{Conclusion}
\label{sec:conclusion}
In this work, we proposed Video Memory Skill Agent~(VMSA), a framework that jointly designs structured video memories and memory skills for long-video question answering. VMSA constructs complementary video memories that serve as searchable evidence sources or provide broad yet detailed video-wide context. It further induces memory skills from successful agent trajectories, enabling the agent to use video-wide context and reusable strategies to guide iterative evidence retrieval, reasoning, and verification. Across seven long-video question-answering benchmarks, VMSA achieves the highest accuracy on all three week-long egocentric benchmarks and the best or competitive performance on four additional benchmarks. Ablation studies further demonstrate the contributions of the memory components and memory skills. These results highlight the importance of jointly designing video memories and mechanisms for their effective utilization in long-video understanding.

\clearpage
\section*{LLM Usage Statement}

We used ChatGPT for sentence-level editing to improve grammar and clarity, Codex to assist with code implementation for this study, and Figma Agent to draw figures. All edits were reviewed and verified by the authors. No scientific ideas, methods, or results were produced by LLMs. The authors reviewed and verified all LLM-assisted text and code and take full responsibility for the content.

{
    \small
    \bibliographystyle{ieeenat_fullname}
    \bibliography{main}
}

\clearpage
\appendix
\section{Dataset Details}
\label{sec:dataset}

We evaluate VMSA on diverse long-video benchmarks, including week-long egocentric benchmarks and other long-video benchmarks. Table~\ref{tab:dataset_details} summarizes the evaluation subsets, including the number of unique source videos or participant streams, the number of evaluation queries, and the average source duration.

\begin{table}[H]
    \centering
    \caption{Benchmark statistics for the datasets used in our experiments. We report unique source videos or participant streams, the number of questions, and the average duration of videos. For the week-long egocentric benchmarks, duration is the total recorded footage per participant stream across seven daily recordings.}
    \label{tab:dataset_details}
    \small
    \setlength{\tabcolsep}{4pt}
    \begin{tabular}{l r r r}
        \toprule
        Dataset & \# Sources & \# Queries & Avg. Dur. \\
        \midrule
        \multicolumn{4}{l}{\emph{Week-long egocentric benchmarks}} \\
        EgoLifeQA & 1 & 500 & 44.3 h \\
        Ego-R1 Bench & 6 & 300 & 44.3 h \\
        EgoMemReason & 6 & 500 & 44.3 h \\
        \midrule
        \multicolumn{4}{l}{\emph{Other long-video benchmarks}} \\
        Video-MME (L) & 300 & 900 & 41.1 min \\
        LongVideoBench (L) & 188 & 564 & 23.9 min \\
        LVBench & 103 & 1,549 & 67.3 min \\
        HippoVlog & 25 & 1,000 & 27.3 min \\
        ScaleLong & 263 & 1,707 & 86.0 min \\
        VideoOdyssey-V & 100 & 1,618 & 109.0 min \\
        \bottomrule
    \end{tabular}
\end{table}

\paragraph{Video-MME.}
Video-MME~\cite{videomme} is a comprehensive multiple-choice benchmark covering six visual
domains and a broad range of perception and reasoning tasks. Its complete
release spans short, medium, and long videos and supports evaluation with or
without subtitle assistance. We evaluate the long-duration subset, comprising
900 questions over 300 videos with an average duration of 41.1 minutes and
report results without subtitles in the main comparison. Results with
subtitles are reported in Table~\ref{tab:appendix_with_subtitle}.

\paragraph{LongVideoBench.}
LongVideoBench~\cite{longvideobench} evaluates long-context, interleaved video-language
understanding through referring reasoning. A question may refer to an earlier
event or statement, requiring the model to identify the referenced context and
then reason over the associated visual and textual evidence. We use the
long-duration portion of the validation split, comprising 564 questions over
188 videos, with an average duration of 23.9 minutes. We report results without
subtitles in the main comparison.

\paragraph{LVBench.}
LVBench~\cite{lvbench} targets extreme long video understanding using videos that range from
roughly 30 minutes to more than two hours. Its questions cover key-information
retrieval, entity recognition, event understanding, temporal grounding,
summarization, and reasoning; a question may belong to more than one category.
The benchmark provides ground-truth interval annotations for temporal
grounding, which we use only in the temporal grounding analysis. We evaluate
1,549 questions over 103 videos.

\paragraph{HippoVlog.}
HippoVlog~\cite{hippomm} evaluates memory formation and associative recall over
continuous audiovisual experiences. It contains 25 long-form daily vlogs and
1,000 human-verified questions, balanced across visual retrieval, auditory
retrieval, cross-modal audiovisual association, and semantic summarization.
Following the setting used by WorldMM~\cite{worldmm}, we evaluate VMSA with
video transcripts. HippoVlog complements visually focused long video QA by
testing whether the agent can preserve and retrieve both modality-specific and
cross-modal evidence.

\paragraph{ScaleLong.}
ScaleLong~\cite{scalelong} evaluates long video understanding at four nested temporal
granularities: clip, shot, event, and story. Questions at all four scales are
constructed over the same video content, which separates temporal-scale
difficulty from differences in video identity or domain. After excluding six corrupted source videos, we evaluate 1,707 questions over 263 valid videos. We report overall accuracy and the per-granularity breakdown in Table~\ref{tab:appendix_scalelong}.

\paragraph{VideoOdyssey-V.}
VideoOdyssey~\cite{videoodyssey} evaluates ultra-long context understanding with videos spanning
11 domains and 54 subcategories. The benchmark distinguishes a visual subset
from an audiovisual subset; we use VideoOdyssey-V to isolate visual
understanding. Our evaluation contains 1,618 questions across 100 videos with an
average duration of approximately 109 minutes. The questions cover perception
tasks such as object, action, attribute, and text recognition, together with
cognition tasks including counting, temporal ordering and grounding, causal 
reasoning, intentional reasoning, spatial reasoning, and summarization.

\paragraph{EgoLifeQA.}
EgoLifeQA~\cite{egolife} is a benchmark constructed from week-long
egocentric recordings captured using wearable glasses by six participants.
It evaluates real-world memory and reasoning capabilities of long-term
egocentric assistants. The benchmark comprises 500 questions focused
on the experiences of Jake, one of the six participants.

\paragraph{Ego-R1 Bench.}
Ego-R1 Bench~\cite{egor1} was introduced with the Ego-R1 framework to evaluate
reasoning-intensive question answering over week-long egocentric video
histories. It contains 300 multiple-choice questions distributed evenly across
six first-person perspectives, with a balanced mixture of human-authored
questions and model-assisted questions that were subsequently verified by
humans. For category-wise evaluation, we follow the WorldMM setting~\cite{worldmm}
and map the original Ego-R1 query types to the EgoLifeQA taxonomy.

\paragraph{EgoMemReason.}
EgoMemReason~\cite{egomemreason} evaluates memory-driven reasoning over week-long EgoLife recordings from six participants. It contains 500 multiple-choice questions generated from video captions and hierarchical event summaries, followed by automatic filtering and human verification of visual grounding, answer correctness, and distractor quality. Answering a question requires integrating an average of 5.1 distinct evidence across segments, with an average temporal range of 25.9 hours.

\section{Implementation Details}
\label{sec:appendix-implementation}

\subsection{Structured Video Memory Construction}
\label{sec:appendix-memory-construction}

We use Qwen3.5-9B as the memory-construction VLM for segment captioning,
local entity profile extraction and merging, Entity Event consolidation,
and Entity-Aware Episode construction. Segment captions, entity profiles,
events, and episodes retain their supporting temporal ranges.

\paragraph{Dense segment captioning.}
For the week-long egocentric benchmarks, we use 60-second segments and
populate Segment Caption Memory with the captions released with
WorldMM~\cite{worldmm}. For all other benchmarks, we divide each video
into 16-second segments with a 12-second stride, yielding a 4-second
overlap between consecutive segments. The memory-construction VLM
generates a caption for each segment from uniformly sampled frames using
a four-shot prompt. We preserve the aspect ratio of each frame and limit
its area to 524{,}288 pixels (0.52~MP), corresponding to an upper bound
of approximately 256 visual tokens per frame.

\paragraph{Local entity profile extraction.}
For each segment, the memory-construction VLM extracts local entity
profiles from uniformly sampled frames. Each profile follows a fixed
schema that records observable attributes, actions, relations, and state
changes together with their supporting temporal range. For the week-long
egocentric benchmarks, the VLM also receives the corresponding segment
caption as input.

\paragraph{Entity profile merging.}
We progressively merge profiles that refer to the same entity within and
across sliding temporal windows. Each window contains up to 32 consecutive
segments. Within a window, we construct a mutual-$k$NN matching over profile
embeddings with $k=6$ to identify candidate groups. We directly merge
high-confidence matches with cosine similarity above 0.85. For the
remaining groups, the memory-construction VLM verifies which profiles
refer to the same entity and merges them by aggregating their attributes.
We repeat this process on the accumulated profiles after every four
windows. After processing the sliding windows, we perform three final video-level merging passes.

\paragraph{Entity Event consolidation.}
For each resulting entity $E_k$, we order its segment-level descriptions
by time. The memory-construction VLM consolidates repeated descriptions
of the same ongoing event and extends its temporal range to cover the
supporting observations. It starts a new event when the observations
indicate a substantive change in the action, participants, relation,
scene, or entity state. Each Entity Event $e_{k,j}$ consists of an event
description and its supporting temporal range.

\paragraph{Entity-Aware Episode construction.}
Within each 1{,}200-second sliding window, we group adjacent segment
captions that describe the same ongoing activity or interaction into
non-overlapping local episodes. For each entity $E_k$, we select candidate
local episodes whose temporal ranges overlap at least one of its Entity
Events. When the candidates span disjoint intervals, the memory-construction
VLM checks each episode for observations of the entity's activities,
states, or interactions. It orders the verified episodes chronologically
and merges those that describe semantically related observations of the
entity. Each resulting entity-aware episode preserves the relevant
participants and interactions and records disjoint supporting intervals
separately. We retain unmerged local episodes as standalone episodes.
Finally, the VLM generates a global video summary from the entity-aware
episodes.

\paragraph{Post-hoc correction of entity profiles.}
After profile merging, Qwen3.6-27B reviews the resulting profiles for
over-merging and under-merging. It corrects over-merging by separating
local profiles that refer to different entities and corrects under-merging
by combining profiles that refer to the same entity.
Section~\ref{sec:appendix-entity-profile-merging-quality} reports a
model-based assessment of profile merging on ScaleLong and VideoOdyssey-V.

\subsection{Model-Based Analysis of Entity Profile Merging}
\label{sec:appendix-entity-profile-merging-quality}

Table~\ref{tab:appendix_entity_merge_quality} reports the reduction in
profile count and the frequency of detected merging errors. On ScaleLong,
profile merging reduces 587{,}736 local entity profiles to 36{,}482
entity records, a 93.8\% reduction. On VideoOdyssey-V, it reduces
217{,}634 local profiles to 18{,}195 entity records, a 91.6\% reduction.

We use Claude Sonnet 5~\cite{claudesonnet5} to identify cases of
over-merging and under-merging. For each error type, we report the number
of detected cases as a percentage of the resulting entity records.
The detected over-merging rate is approximately 0.2\% on both benchmarks.
The detected under-merging rate is 0.6\% on ScaleLong and 0.7\% on
VideoOdyssey-V.

\begin{table}[H]
  \centering
  \caption{Entity canonicalization statistics and model-based identity audit on
           ScaleLong and VideoOdyssey-V. \emph{Raw Entity} counts
           segment-level entity profiles, whereas
           \emph{Canonical Entity} counts the resulting merged entities.
           \emph{Over-merged} and \emph{Under-merged} report cases identified
           by Claude Sonnet 5.}
  \label{tab:appendix_entity_merge_quality}
  \small
  \setlength{\tabcolsep}{4pt}
  \resizebox{\columnwidth}{!}{%
  \begin{tabular}{l rrrr}
    \toprule
    Dataset & Raw Entity & Canonical Entity & Over-merged & Under-merged \\
    \midrule
    ScaleLong      & 587{,}736 & 36{,}482 & 73 & 204 \\
    VideoOdyssey-V & 217{,}634 & 18{,}195 & 36 & 122 \\
    \bottomrule
  \end{tabular}
  }
\end{table}

\subsection{Search and Verification Tool Details}
\label{sec:appendix-memory-search}

\begin{figure*}[!t]
  \input{tab/appendix_agent_prompts}
\end{figure*}

This section details the tools summarized in Section~\ref{sec:agentic-search}.
Here, $\operatorname{Enc}(\cdot)$ denotes the embedding model, which supports
both multimodal and text-only inputs.

\paragraph{Caption Search.}
Caption Search retrieves segment captions and their temporal ranges from
Segment Caption Memory $\mathcal{M}_{\mathrm{cap}}$. Based on the question
and accumulated evidence, the agent formulates a query $q_{\mathrm{cap}}$
that specifies an activity, event, or visual cue to locate. The tool ranks
segments by cosine similarity and returns the $K_c$ highest-ranked
caption--range pairs:
\begin{equation}
    \mathcal{R}_{\mathrm{cap}}(q_{\mathrm{cap}})
    = \underset{(C_i,\tau_i)\in\mathcal{M}_{\mathrm{cap}}}
      {\operatorname{TopK}_{K_c}}
      \; \cos\!\left(\operatorname{Enc}(q_{\mathrm{cap}}), z_i\right),
    \label{eq:caption-search}
\end{equation}
where $z_i$ is the precomputed embedding of segment $s_i$, obtained by
encoding its caption $C_i$ either alone or together with frames $F_i$
from the segment. Each retrieved caption $C_i$ provides a fine-grained description of the segment, and its temporal
range $\tau_i$ identifies a candidate interval for further inspection.

\paragraph{Entity Event Search.}
Entity Event Search retrieves events involving the same entity across
potentially distant intervals in Entity Event Memory
$\mathcal{M}_{\mathrm{evt}}$. Based on the question and accumulated
evidence, the agent formulates a query $q_{\mathrm{evt}}$ that specifies
the target entity and a relevant action, relation, or state. The tool
selects the $K_e$ events with the highest cosine similarity to the query:
\begin{equation}
    \mathcal{R}_{\mathrm{evt}}(q_{\mathrm{evt}})
    = \underset{1 \leq k \leq N_e,\; 1 \leq j \leq J_k}
      {\operatorname{TopK}_{K_e}}
      \; \cos\!\left(
        \operatorname{Enc}(q_{\mathrm{evt}}),
        \operatorname{Enc}(e_{k,j})
      \right),
    \label{eq:entity-event-search}
\end{equation}
where $e_{k,j}$ denotes the $j$-th event associated with entity $E_k$.
For each selected event, the tool returns its description, its supporting
temporal range, and the profile of the associated entity $E_k$.

\paragraph{Segment Inspection.}
Segment Inspection examines source-video frames when textual evidence is
insufficient to distinguish among candidate answers. The agent uses
Entity-Aware Episode Memory in its initial context or results from Caption
Search or Entity Event Search to identify relevant temporal intervals.
Before inspecting the selected intervals, the agent formulates a
verification query $q_t^{\mathrm{ver}}$ that specifies the visual details
needed to distinguish among the remaining interpretations. For example,
the query may ask who performs an action, what state an object is in, or
which of two events occurs first. It may also target object identity,
poses or gestures, on-screen text, instance counts, or spatial relations.

Let $\{s_{i_1},\ldots,s_{i_m}\}$ denote the selected segments and
$\mathcal{F}_i$ the set of frames sampled uniformly from segment $s_i$.
Given the verification query and the sampled frames, the agent model
returns an observation
\begin{equation}
    o_t^{\mathrm{ins}}
    = \Phi_{\mathrm{ins}}\!\left(
        q_t^{\mathrm{ver}},
        \{\mathcal{F}_{i_1},\ldots,\mathcal{F}_{i_m}\}
      \right),
    \label{eq:segment-inspection}
\end{equation}
where $\Phi_{\mathrm{ins}}$ denotes the agent model used for inspection.
The observation $o_t^{\mathrm{ins}}$ reports what is visible in the
sampled frames and indicates which candidate interpretations this evidence
supports or contradicts. The agent adds the observation to its accumulated
evidence and, if needed, inspects an adjacent interval or another relevant
part of the video.

\begin{figure*}[!t]
  \centering
  \begin{paperartifactbox}{Prompt Template for Segment Inspection}
    \footnotesize\ttfamily
    Carefully inspect the source-video frames sampled from the selected
    segment spans. Pay attention to the cause and sequence of events, the
    detail and movement of objects, and the action and pose of persons.\par
    \vspace{0.35em}
    Question: \{question\}\par
    Candidates: \{choices\}\par
    Selected segment spans: \{selected\_spans\}\par
    Verification points: \{verification\_points\}\par
    \vspace{0.35em}
    First describe what is directly visible in this span. Do not choose an
    answer yet.\par
    \textbf{Output 1: Span observations.} Write 3--6 short bullets. Describe
    only what is directly visible in this span and provide details that
    discriminate among candidates. Include exact action, object state, person
    pose or gesture, screen text, diagram, table, icon, count, color, spatial
    side, or temporal order when visible. Describe the state before, at, and
    after the anchor when visible.\par
    \textbf{Output 2: Verification table.} For each verification point, copy
    the point and report direct evidence in this span, option clauses directly
    supported, option clauses contradicted, and the exact missing fact plus
    whether it needs a closer frame, adjacent span, subtitle or audio anchor,
    or another candidate segment.
  \end{paperartifactbox}
  \captionof{figure}{\textbf{Prompt used for Segment Inspection.} The agent
  model supplies bounded temporal spans and answer-discriminative verification
  points. The inspection model reports directly observed evidence from the
  corresponding source-video frames.}
  \label{fig:appendix_inspection_prompt}
\end{figure*}

\paragraph{Tool configuration.}
On the week-long egocentric benchmarks, it retrieves 15 segment captions per call using caption embeddings. On the
other long-video benchmarks, it retrieves 10 segment captions per call using multimodal embeddings of video-caption pairs. Across all benchmarks, Entity Event Search uses the agent-generated query to retrieve 20 Entity Events per call. For each event, it returns the event description, its supporting temporal range, and the associated entity profile. We disable Segment Inspection on the week-long egocentric benchmarks to reduce inference costs and enable it on the other long-video benchmarks. Unless otherwise specified, Segment Inspection uses the same model as the reasoning agent.

\subsection{Memory-Skill Induction}
\label{sec:appendix-skill-induction}

\paragraph{Memory-skill structure.}
Each memory skill names a procedure and specifies the condition under which
the agent should apply it. The skill provides ordered steps for retrieving
and verifying evidence, a checklist of intermediate findings to retain
across search turns, and a reasoning template for interpreting the
accumulated evidence. It also specifies when the agent has sufficient
evidence to stop searching and identifies common errors to avoid.
Figure~\ref{fig:appendix_skill_example} shows the memory-skill template
and an example of an induced skill.

\paragraph{Collecting successful trajectories.}
For each library, we collect trajectories from source benchmarks distinct
from its evaluation benchmarks. We run the Qwen3.6-27B agent with
structured video memory and the search and verification tools, without
memory-skill guidance. We retain trajectories for which the agent's final
prediction is correct. To reduce the context length during induction, we
omit raw tool results while preserving the agent's reasoning, tool calls,
and the temporal ranges of the evidence it uses.

\paragraph{Incremental library update and consolidation.}
We use GPT-5.6 Sol~\cite{gpt56} as the primary induction LLM. Starting
from an empty library, the model processes successful trajectories
sequentially. For each trajectory, it assesses whether the current skills
cover the retrieval, reasoning, and verification procedures demonstrated
by the agent. If coverage is sufficient, the model leaves the library
unchanged. Otherwise, it refines an existing skill to capture the missing
procedure or adds a new skill if the procedure falls outside the scope of
the existing skills. In both cases, the model abstracts away details
specific to the question or video.

After every 100 trajectories, the induction LLM consolidates the library.
It merges skills only when their trigger conditions, retrieval and
verification steps, evidence-retention requirements, and stopping rules
describe the same procedure. During merging, the model preserves the
procedural guidance from each constituent skill.

\paragraph{Cross-benchmark memory-skill transfer.}
We construct three libraries. For evaluation on Video-MME (L),
LongVideoBench (L), LVBench, and HippoVlog, we use a 15-skill library
induced from 2{,}181 successful trajectories on ScaleLong and
VideoOdyssey-V~\cite{scalelong,videoodyssey}. For evaluation on ScaleLong
and VideoOdyssey-V, we use a 19-skill library induced from 2{,}549
successful trajectories on Video-MME (L), LongVideoBench (L), LVBench,
and EgoSchema~\cite{egoschema}. For the three week-long egocentric
benchmarks (EgoLifeQA, Ego-R1 Bench, and EgoMemReason), we use a 26-skill
library induced from trajectories on six source benchmarks: ScaleLong,
VideoOdyssey-V, Video-MME, LongVideoBench, LVBench, and EgoSchema.

\begin{figure*}[t]
  \centering
  \begin{paperartifactbox}{Memory-Skill Template and Example}
    \footnotesize
    \textbf{Memory-skill template.}
    Each retrieved memory skill contains a \texttt{procedure\_name}, \texttt{trigger},
    ordered \texttt{steps}, an \texttt{evidence\_memory} checklist, a
    \texttt{reasoning\_template}, a \texttt{stop\_rule}, and an
    \texttt{avoid} list. A representative memory skill is shown below.

    \vspace{0.5em}\hrule\vspace{0.5em}
    \textbf{TopicIntervalLocalization} \emph{(temporal localization)}

    \begin{minipage}[t]{0.48\linewidth}
      \textbf{Trigger.} Use when asked for the start and end time of a topic,
      event, or discussion whose core material must be localized within a
      segmented long video.

      \textbf{Steps.}
      \begin{enumerate}[leftmargin=*,nosep]
        \item Expand the target into exact terms, abbreviations, synonyms,
              precursor events, characteristic imagery, and closing events.
        \item Search from specific phrases to broader associated cues until at
              least one high-confidence core segment is found.
        \item Use Segment Inspection to move backward from the hit to the first
              segment substantially about the target and forward to the last
              contiguous core segment before the topic changes.
        \item Classify boundary evidence as introduction, core discussion,
              continuation, incidental mention, or transition so peripheral
              references do not widen the interval.
        \item Bracket both sides with neighboring non-target content and
              preserve uncertainty when the topic begins or ends within a
              segment.
        \item Convert segment bounds to clock time using the actual segment
              duration, indexing origin, and end-bound convention, and check
              the arithmetic.
        \item Compare candidate ranges by overlap with the verified core span
              and reject ranges centered on precursor or adjacent topics.
      \end{enumerate}

      \textbf{Stop rule.} Stop when both topic boundaries are bracketed by
      neighboring non-target content, the segment-to-time mapping and
      arithmetic are checked, boundary uncertainty is recorded, and only one
      candidate range adequately covers the contiguous core interval.
    \end{minipage}
    \hfill
    \begin{minipage}[t]{0.49\linewidth}
      \textbf{Evidence memory.}
      \begin{itemize}[leftmargin=*,nosep]
        \item Search vocabulary grouped as exact term, abbreviation, synonym,
              precursor, characteristic visual cue, and closing cue.
        \item Verified first core segment, last core segment, and neighboring
              non-target segments that bracket the interval.
        \item Classification of each inspected segment as introduction, core
              discussion, continuation, incidental mention, or transition.
        \item Segment duration, indexing origin, start-bound formula,
              end-bound formula, and partial-segment uncertainty.
        \item Overlap of each candidate range with the verified core interval.
      \end{itemize}

      \textbf{Reasoning template.}
      \begin{enumerate}[leftmargin=*,nosep]
        \item Searches locate the target's core material near
              \texttt{<candidate region>}; backward and forward Segment Inspection
              calls establish contiguous bounds \texttt{<first>} through
              \texttt{<last>}.
        \item Neighboring segments contain \texttt{<non-target content>}, so
              precursor and retrospective mentions are excluded from the core
              interval.
        \item Using segment duration \texttt{<duration>} and indexing origin
              \texttt{<origin>}, convert the verified bounds with
              \texttt{<formulas>}; select the range that covers the core span
              and excludes adjacent topics.
      \end{enumerate}

      \textbf{Avoid.} Do not stop after one failed exact-phrase search; equate
      a precursor, incidental mention, or retrospective reference with the
      contiguous core interval; mix one-based labels with zero-based time
      calculations; report precise clock bounds without checking duration,
      indexing origin, end-bound convention, and partial-segment uncertainty;
      or choose a range merely because it touches the topic while being
      centered on adjacent content.
    \end{minipage}
  \end{paperartifactbox}
  \caption{\textbf{Memory-skill schema and representative example.} Each
  memory skill specifies an activation condition, ordered evidence-seeking
  steps, an evidence-memory checklist, a reasoning template, a stopping rule,
  and common failure modes.}
  \label{fig:appendix_skill_example}
\end{figure*}

\subsection{Memory-Skill Induction Prompts}
\label{sec:appendix-skill-prompts}

Figure~\ref{fig:appendix_skill_induction_prompts} shows abridged prompts
for trajectory coverage assessment and skill library consolidation.
During coverage assessment, the induction LLM checks whether the current
library covers the retrieval, reasoning, and verification procedures
demonstrated in a successful trajectory. If a procedure is missing, the
model proposes a revision to an existing skill or adds a proposal for a
new skill. Each proposal includes the complete skill record and identifies
the supporting trajectories. During consolidation, the model applies
compatible revisions, incorporates distinct new skills, and merges
redundant proposals. Both prompts use the skill template in
Figure~\ref{fig:appendix_skill_example}. For readability, the figure
omits bookkeeping and serialization details.

\begin{figure*}[!tbp]
  \centering
  \footnotesize
  \begin{minipage}[t]{0.60\linewidth}
    \vspace{0pt}
    \begin{paperartifactbox}{Skill Induction}
      \raggedright
      \textbf{Input.} The current skill library and successful source
      trajectories.\par
      \medskip
      Assess each trajectory against the current library. Consider it covered
      only if existing skills specify its information-need decomposition,
      memory and tool selection, evidence retention, reasoning or verification,
      and stopping decisions.\par
      \smallskip
      For covered trajectories, identify the skills that provide coverage.
      For an uncovered
      trajectory, explain the missing procedural knowledge and propose either
      a revision of a compatible skill or a new skill for a distinct
      procedure. Use the shared skill template for both operations.\par
      \smallskip
      Abstract reusable procedures. Exclude video-specific facts, entities,
      answers, answer choices, and trajectory identifiers from skill
      descriptions.\par
      \medskip
      \textbf{Output.} A coverage assessment for each trajectory and proposed
      library changes. Each proposal identifies the supporting trajectories
      and provides either a complete revision of an existing skill or a
      complete new skill record.
    \end{paperartifactbox}
  \end{minipage}
  \hfill
  \begin{minipage}[t]{0.38\linewidth}
    \vspace{0pt}
    \begin{paperartifactbox}{Library Consolidation}
      \raggedright
      \textbf{Input.} The current skill library after incremental
      induction.\par
      \medskip
      Merge skills only when their trigger conditions, retrieval and
      verification steps, evidence-retention requirements, and stopping
      criteria describe the same underlying procedure.\par
      \smallskip
      Preserve their reusable procedural knowledge when merging, and retain
      skills that represent distinct procedures.\par
      \medskip
      \textbf{Output.} A consolidated skill library with redundant skills
      merged while preserving their reusable procedural knowledge.
    \end{paperartifactbox}
  \end{minipage}
  \caption{\textbf{Abridged prompts for memory-skill induction and library
  consolidation.} Induction produces coverage assessments and complete revised
  or new skill records. Consolidation merges skills representing the same
  underlying procedure while preserving their reusable procedural knowledge.}
  \label{fig:appendix_skill_induction_prompts}
\end{figure*}

\subsection{Catalog of Memory Skills}
\label{sec:appendix-skill-catalog}

Table~\ref{tab:appendix_skill_catalog} lists the 15 memory skills induced
from successful trajectories on ScaleLong and VideoOdyssey-V and used
for evaluation on Video-MME (L), LongVideoBench (L), LVBench, and
HippoVlog (Section~\ref{sec:appendix-skill-induction}). The skills guide
the agent in determining whether an event meets the conditions in the
question, reconstructing the order of events, defining the scope of a
search, and comparing observations. Each skill specifies when its procedure
applies, which evidence to retrieve and retain, how to verify unresolved
claims, and when sufficient evidence has been gathered to stop the search.

\begin{table}[!htbp]
  \centering
  \caption{Catalog of the 15 memory skills transferred to the four conventional
           long-video benchmarks. Each entry summarizes the evidence-seeking
           procedure encoded by a skill.}
  \label{tab:appendix_skill_catalog}
  \footnotesize
  \setlength{\tabcolsep}{4pt}
  \renewcommand{\arraystretch}{1.12}
  \begin{tabular}{@{}>{\raggedright\arraybackslash}p{0.44\columnwidth}%
                     >{\raggedright\arraybackslash}p{\dimexpr0.56\columnwidth-8pt\relax}@{}}
    \toprule
    Skill name & Evidence-seeking procedure \\
    \midrule
    Constraint Driven Multimodal Evidence Resolution &
      Decompose constraints, track unresolved distinctions, and verify each
      component and its relationship. \\
    Evidence Anchored Semantic Claim Discrimination &
      Test competing claims clause by clause, including attribution, outcomes,
      quantifiers, and causal links. \\
    Temporal Event Ledger Reconstruction &
      Reconstruct event order, boundaries, and elapsed time while distinguishing
      attempts, continuations, and replays. \\
    Boundary Aware Unit Enumeration &
      Define unit boundaries and hierarchy before counting repeated or nested
      events. \\
    Interval Prevalence Comparison &
      Compare counts, exposure durations, or rates under consistent scope and
      duplicate rules. \\
    Topic Interval Localization &
      Locate core topic content and bracket its boundaries with neighboring
      non-target evidence. \\
    Observer Relative Spatial Relation Resolution &
      Ground the scene layout and observer pose before resolving directions
      or route steps. \\
    Scope Bounded Topic Synthesis &
      Integrate representative in-scope evidence while separating central
      themes from incidental content. \\
    Frame Bounded Visible Participant Enumeration &
      Count visible participants within an instant or short window while
      reconciling occlusion and repeated views. \\
    Synchronized Clock State Extraction &
      Align an embedded clock with video time and read the associated state
      from the same display. \\
    Frame Bounded Visible Object Enumeration &
      Define countable objects and reconcile grouping ambiguity, occlusion,
      and repeated views. \\
    Evidence Normalized Comparative Potential Assessment &
      Compare candidates using common performance criteria, observation
      scope, and supported career-stage evidence. \\ \\
    Evidence Calibrated Comparative Risk Assessment &
      Compare hazards, exposure, safeguards, and uncertainty under a
      consistent risk rubric. \\
    Recurring Entity Appearance Boundary Localization &
      Track a recurring entity and refine its first and last on-screen
      appearance boundaries. \\
    Entity Grounded External Attribute Resolution &
      Ground an entity or event instance before resolving external facts,
      keeping them separate from video evidence. \\
    \bottomrule
  \end{tabular}
\end{table}

\subsection{Model and Decoding Configuration}

We run Qwen3.5-9B and Qwen3.6-27B in instruct mode with thinking disabled.
Unless otherwise specified, decoding uses a temperature of 0.7, top-$p$ of
0.8, top-$k$ of 20, presence penalty of 1.5, and repetition penalty of 1.0.
We use Qwen3-VL-Embedding-2B~\cite{qwen3vlembedding} for both multimodal and text retrieval
embeddings.

\section{Evidence Retrieval and Temporal Grounding}
\label{sec:appendix-temporal-grounding}

We evaluate the temporal grounding of VMSA's search using temporal
Intersection-over-Ground Truth (tIoG) and temporal Intersection-over-Union
(tIoU). For each tool turn, we compare the union of retrieved temporal ranges
with the union of ground-truth intervals. tIoG divides the duration of their intersection by the total ground-truth duration and measures evidence coverage. tIoU divides the same intersection duration by the duration of their union, measuring overlap between the retrieved and ground-truth intervals. We use these metrics to examine whether
better-grounded trajectories are associated with more accurate answers.

\subsection{Temporal Grounding and Answer Accuracy}

We examine the association between evidence coverage and answer accuracy by grouping trajectories according to whether their best-turn tIoG exceeds 0.5 (Table~\ref{tab:analysis_grounding_accuracy}). 
The group with tIoG greater than 0.5 achieves answer accuracy that is 3.6\%, 9.1\%, and 13.1\% higher on LVBench, ScaleLong, and VideoOdyssey-V, respectively.

\begin{table}[H]
  \centering
  \caption{Association between temporal grounding and answer accuracy. For
           each trajectory, we use its best-turn tIoG and partition trajectories
           at 0.5. We report answer accuracy (\%) for each group; $\Delta$
           denotes their difference.}
  \label{tab:analysis_grounding_accuracy}
  \small
  \begin{tabular*}{\columnwidth}{@{\extracolsep{\fill}}lccc@{}}
    \toprule
    Dataset & tIoG $\leq 0.5$ & tIoG $> 0.5$ & $\Delta$ \\
    \midrule
    LVBench        & 71.0 & \textbf{74.6} & +3.6 \\
    ScaleLong      & 65.1 & \textbf{74.2} & +9.1 \\
    VideoOdyssey-V & 57.8 & \textbf{70.9} & +13.1 \\
    \bottomrule
  \end{tabular*}
\end{table}

\subsection{Evidence Attribution Across Memory Views}
\label{sec:appendix-memory-credit}

Table~\ref{tab:appendix_memory_utilization} reports evidence credit for each
memory view over correct search trajectories. We use Grok-4.6 to assess the records provided by each memory view independently, given the question, candidate answers, and ground-truth answer. A memory view receives credit if it explicitly states a fact that favors the correct answer over the distractors or if its temporal support overlaps any segment examined during Segment Inspection. Entity-Aware Episodes receive the highest credit on both benchmarks. Segment Captions receive credit more frequently than Entity Events on Video-MME (L), whereas the reverse holds on HippoVlog, indicating that the more frequently credited local evidence source varies by dataset.

\begin{table}[H]
  \centering
  \caption{Evidence credit (\%) of structured-memory views over correct
           Video-MME (L) and HippoVlog trajectories, conditioned
           on each view providing records. A view receives credit when its
           records directly state an answer-discriminative fact or their
           temporal support overlaps segments examined by Segment Inspection.
           Views are credited independently.}
  \label{tab:appendix_memory_utilization}
  \small
  \setlength{\tabcolsep}{6pt}
  \begin{tabular}{@{}l ccc@{}}
    \toprule
    Dataset & Episodes & Captions & Entity Events \\
    \midrule
    Video-MME (L) & 88.3 & 84.3 & 68.4 \\
    HippoVlog     & 97.6 & 74.5 & 79.9 \\
    \bottomrule
  \end{tabular}
\end{table}

\section{Robustness and Scaling}

\subsection{Effect of Entity-Aware Episodes}
\label{sec:appendix-episode-context}

Table~\ref{tab:appendix_entity_episodes_ablation} evaluates the contribution
of entity-aware episodes to the agent's initial context. We compare two
configurations: one that provides only the global video summary $G$ and one
that also provides the entity-aware episodes. Both use the same Qwen3.6-27B
agent, memory skills, search and verification tools, and interaction budget.
Adding entity-aware episodes improves answer accuracy by 4.3\% on Video-MME (L), 2.9\% on LVBench, and 5.7\% on HippoVlog.

\begin{table}[!htbp]
  \centering
  \caption{Effect of Entity-Aware Episodes in the initial context. $G$ denotes
           the global video summary, and Episode denotes Entity-aware Episode Memory. We report
           multiple-choice accuracy (\%). }
  \label{tab:appendix_entity_episodes_ablation}
  \setlength{\tabcolsep}{3pt}
  \begin{tabular*}{\columnwidth}{@{\extracolsep{\fill}}ccccc@{}}
    \toprule
    $G$ & Episode & \shortstack{Video-\\MME (L)} & LVBench & HippoVlog \\
    \midrule
    \checkmark & -          & 71.8 & 69.9 & 76.4 \\
    \checkmark & \checkmark & 76.1 & 72.8 & 82.1 \\
    \bottomrule
  \end{tabular*}
\end{table}

\subsection{Extended Evaluation of Memory Skills}
\label{sec:appendix-memory-skills}

Table~\ref{tab:appendix_ablation_memory_skills} extends the memory-skill
ablation to three benchmarks without ground-truth temporal evidence
intervals. We therefore evaluate answer accuracy, comparing VMSA with and
without memory skills while holding the Qwen3.6-27B agent, Entity-Aware
Episode Memory, Caption Search, and Entity Event Search fixed. Memory skills
improve accuracy on all three benchmarks, from 74.6\% to 76.1\% on
Video-MME (L), from 71.3\% to 74.1\% on LongVideoBench (L), and from
81.4\% to 82.1\% on HippoVlog.

\begin{table}[!htbp]
  \centering
  \caption{Effect of memory skills on answer accuracy (\%) with the
           Qwen3.6-27B agent.}
  \label{tab:appendix_ablation_memory_skills}
  \setlength{\tabcolsep}{3pt}
  \begin{tabular}{l c c}
    \toprule
    Dataset & w/o Skills & w/ Skills \\
    \midrule
    Video-MME (L)     & 74.6 & 76.1 \\
    LongVideoBench (L) & 71.3 & 74.1 \\
    HippoVlog          & 81.4 & 82.1 \\
    \bottomrule
  \end{tabular}%
\end{table}

\subsection{Effect of Segment Duration}
\label{sec:ana_segment_scale}

Table~\ref{tab:analysis_segment_size} compares segment durations of 16, 30,
and 60 seconds for structured video memory construction using the same
Qwen3.6-27B agent for evaluation. The default duration of 16 seconds yields
the highest answer accuracy on both benchmarks, reaching 72.8\% on LVBench
and 74.1\% on LongVideoBench (L). Across the two benchmarks, the 30- and
60-second settings reduce accuracy by 4.1 to 5.7\% relative
to the default.

On both benchmarks, most of the decrease occurs when segment duration
increases from 16 to 30 seconds, with a smaller decline at 60 seconds.
Longer segments can group multiple activities or state changes within a
single temporal interval, which may make it harder for Caption Search and
Entity Event Search to localize the evidence needed to answer a question.

\begin{table}[H]
  \centering
  \caption{Effect of segment duration on structured video memory. We vary the segment duration from 16 to 60\,s and
           report accuracy (\%) on LVBench and LongVideoBench (L). The 16\,s
           setting achieves the best accuracy on both benchmarks; longer
           segments yield fewer but temporally coarser memory records.}
  \label{tab:analysis_segment_size}
  \small
  \setlength{\tabcolsep}{7pt}
  \begin{tabular}{c cc}
    \toprule
    Segment duration & LVBench & LongVideoBench (L) \\
    \midrule
    16 s (default) & \textbf{72.8} & \textbf{74.1} \\
    30 s           & 68.7          & 68.6 \\
    60 s           & 68.2          & 68.4 \\
    \bottomrule
  \end{tabular}
\end{table}

\subsection{Effect of the Memory-Skill Induction Model}

Table~\ref{tab:appendix_skill_scaling} compares memory skills induced by
GPT-5.6 and Qwen3.6-27B. We use the same induction protocol for both
models and transfer each resulting library to the evaluation benchmarks
without modification. We hold the Qwen3.6-27B reasoning agent, structured
video memory, retrieval tools, and interaction budget fixed.

The two libraries yield the same answer accuracy on Ego-R1 Bench. On
HippoVlog and EgoLifeQA, the agent achieves higher accuracy with the
GPT-5.6 library, with gains of 1.0\% and 2.8\%,
respectively, over the Qwen3.6-27B library.

\begin{table}[H]
  \centering
  \caption{Effect of the memory-skill induction model on HippoVlog,
           EgoLifeQA, and Ego-R1 Bench. We report multiple-choice accuracy (\%) while holding
           the Qwen3.6-27B reasoning agent, structured memory, and inference setting fixed. VMSA achieves comparable performance with memory skills induced by Qwen3.6-27B on Ego-R1 Bench. }
  \label{tab:appendix_skill_scaling}
  \setlength{\tabcolsep}{3.5pt}
  \resizebox{\columnwidth}{!}{%
  \begin{tabular}{l ccc}
    \toprule
    Induction Model & HippoVlog & EgoLifeQA & Ego-R1 Bench \\
    \midrule
    GPT-5.6       & \textbf{82.1} & \textbf{66.8} & \textbf{74.7} \\
    Qwen3.6-27B   & 81.1          & 64.0          & \textbf{74.7} \\
    \bottomrule
  \end{tabular}%
  }
\end{table}

\subsection{Extended Agent-Model Comparison}

Table~\ref{tab:appendix_full_table_scaling} extends the agent-model
comparison to LongVideoBench (L), LVBench, and HippoVlog. We vary the
agent model while holding structured video memory and inference settings
fixed. Replacing Qwen3.5-9B with Qwen3.5-122B-A10B improves answer
accuracy by 9.4\%, 7.9\%, and 2.0\% on the three benchmarks,
respectively. Switching from Qwen3.5-122B-A10B to the default Qwen3.6-27B
yields further gains of 5.8\%, 2.7\%, and 0.1\%. The three models thus
follow the same accuracy ordering across benchmarks, although the gap
between Qwen3.5-9B and Qwen3.6-27B varies from 2.1\% on HippoVlog
to 15.2\% on LongVideoBench (L).

\begin{table*}[!tbp]
  \centering
  \caption{Full agent-model scaling comparison with the same preconstructed
           video memory. We report accuracy (\%) on four hour-scale
           long video and three week-long egocentric benchmarks. Best results are \textbf{bold}. }
  \label{tab:appendix_full_table_scaling}
  \setlength{\tabcolsep}{4pt}
  \resizebox{\linewidth}{!}{%
  \begin{tabular}{l ccccccc}
    \toprule
    Agent Model & Video-MME (L) & LongVideoBench (L) & LVBench & HippoVlog & EgoLifeQA & EgoMemReason & Ego-R1 Bench \\
    \midrule
    Qwen3.5-9B        & 66.9          & 58.9          & 62.2          & 80.0          & 55.8          & 35.4          & 65.7          \\
    Qwen3.5-122B-A10B & 73.4          & 68.3          & 70.1          & 82.0          & 62.8          & 38.4          & 71.3          \\
    Qwen3.6-27B       & \textbf{76.1} & \textbf{74.1} & \textbf{72.8} & \textbf{82.1} & \textbf{66.8} & \textbf{38.6} & \textbf{74.7} \\
    \bottomrule
  \end{tabular}%
  }
\end{table*}

\subsection{Run-to-Run Variability}
\label{sec:appendix-run-variance}

Across three runs on LVBench under the same configuration, VMSA achieves a
mean answer accuracy of 72.51\% with a standard deviation of 0.27\%.

\section{Detailed Benchmark Results}
\label{sec:appendix-additional-benchmarks}

\subsection{Category-Wise Results on Week-Long Egocentric Benchmarks}
\label{sec:Category-Wise Results on Week-Long Egocentric Benchmarks}

Table~\ref{tab:appendix_egolife_category} breaks down the overall results
in Table~\ref{tab:egolifeqa_egor1} by question category, using the baseline
results reported by WorldMM~\cite{worldmm}. With Qwen3.6-27B as the agent
model, VMSA achieves higher overall accuracy than all compared baselines
on both EgoLifeQA and Ego-R1 Bench. On EgoLifeQA, VMSA improves over
WorldMM by 3.3\% on HabitInsight, 4.0\% on
RelationMap, and 3.2\% on TaskMaster. On Ego-R1 Bench, the largest
gains over WorldMM are on TaskMaster (13.1\%) and EntityLog
(12.6\%). Using GPT-5 as the agent model further increases overall
accuracy to 70.0\% on EgoLifeQA and 75.0\% on Ego-R1 Bench.

\begin{table*}[!tbp]
    \centering
    \caption{Category-wise accuracy (\%) on EgoLifeQA and Ego-R1 Bench. We include baselines from Table~\ref{tab:egolifeqa_egor1} with category-wise results reported by WorldMM~\cite{worldmm}. Ent., EvR., Hab., Rel., and Task denote EntityLog, EventRecall, HabitInsight, RelationMap, and TaskMaster, respectively. The best and second-best results are shown in \textbf{bold} and \underline{underlined}, respectively.}
    \label{tab:appendix_egolife_category}
    \small
    \setlength{\tabcolsep}{3.6pt}
    \resizebox{\textwidth}{!}{%
    \begin{tabular}{l cccccc cccccc}
        \toprule
        & \multicolumn{6}{c}{EgoLifeQA} & \multicolumn{6}{c}{Ego-R1 Bench} \\
        \cmidrule(lr){2-7} \cmidrule(lr){8-13}
        Method
        & Ent. & EvR. & Hab. & Rel. & Task & Overall
        & Ent. & EvR. & Hab. & Rel. & Task & Overall \\
        \midrule
        \multicolumn{13}{l}{\textit{Base Models}} \\
        Gemini-2.5-Pro~\cite{gemini25}
        & 43.2 & 40.5 & 41.0 & 55.2 & 52.4 & 46.4
        & 43.9 & 56.1 & 53.9 & 47.4 & 47.4 & 46.7 \\
        GPT-5~\cite{gpt5}
        & 47.2 & 42.1 & 47.5 & 53.6 & 55.6 & 48.6
        & 41.8 & 58.5 & 53.9 & 52.6 & 50.0 & 46.3 \\
        \midrule
        \multicolumn{13}{l}{\textit{Long-Video LLMs}} \\
        VideoChat-Flash~\cite{videochatflash}
        & 28.8 & 32.5 & 37.7 & 37.6 & 38.1 & 34.2
        & 43.4 & 43.9 & 38.5 & 31.6 & 44.7 & 42.7 \\
        Time-R1~\cite{timer1}
        & 39.2 & 50.8 & 65.6 & 48.8 & 47.6 & 48.8
        & 49.2 & 48.8 & 46.2 & 42.1 & 44.7 & 48.0 \\
        Video-RTS~\cite{videorts}
        & 40.8 & 48.4 & 62.3 & 48.8 & 47.6 & 48.2
        & 47.6 & 46.3 & 53.9 & 52.6 & 47.4 & 48.0 \\
        \midrule
        \multicolumn{13}{l}{\textit{Retrieval-Augmented Methods}} \\
        LightRAG~\cite{lightrag}
        & 40.8 & 48.4 & 67.2 & 50.4 & 44.4 & 48.8
        & 54.0 & 61.0 & 46.2 & 42.1 & 42.1 & 52.3 \\
        HippoRAG~\cite{hipporag}
        & 48.8 & 60.3 & 70.5 & 60.8 & 66.7 & 59.6
        & 54.5 & \underline{65.9} & 69.2 & 52.6 & 50.0 & 56.0 \\
        Video-RAG~\cite{videorag_aligned}
        & 49.6 & 56.3 & 67.2 & 55.2 & 54.0 & 55.4
        & 48.7 & 58.5 & 53.9 & 47.4 & 44.7 & 49.7 \\
        EgoRAG~\cite{egolife}
        & 40.0 & 56.3 & 62.3 & 54.4 & 52.4 & 52.0
        & 46.6 & 56.1 & 46.2 & 47.4 & 55.3 & 49.0 \\
        HippoMM~\cite{hippomm}
        & 45.6 & 53.2 & 70.5 & 55.2 & 58.7 & 54.6
        & 51.9 & 56.1 & 46.2 & 52.6 & 57.9 & 53.0 \\
        \midrule
        \multicolumn{13}{l}{\textit{Agentic Video Frameworks}} \\
        Ego-R1~\cite{egor1} (3B)
        & 51.2 & 53.2 & 63.9 & 50.4 & 50.8 & 53.0
        & 50.8 & 63.4 & 38.5 & 36.8 & 57.9 & 52.0 \\
        M3-Agent~\cite{m3agent} (7B)
        & 44.4 & 54.8 & 62.3 & 56.8 & 54.0 & 53.5
        & 52.4 & 58.5 & 38.5 & 42.1 & 52.6 & 52.0 \\
        WorldMM~\cite{worldmm} (GPT-5)
        & \textbf{62.4} & \underline{64.3} & \underline{75.4} & 62.4 & 71.4 & 65.6
        & 64.6 & \textbf{70.7} & \underline{76.9} & \underline{57.9} & 63.2 & 65.3 \\
        \midrule
        \textbf{VMSA} (Qwen3.6-27B)
        & \underline{61.6} & 62.7 & \textbf{78.7} & \underline{66.4} & \underline{74.6} & \underline{66.8}
        & \underline{77.2} & \underline{65.9} & \underline{76.9} & \textbf{63.2} & \textbf{76.3} & \underline{74.7} \\
        \textbf{VMSA} (GPT-5)
        & \underline{61.6} & \textbf{66.7} & \underline{75.4} & \textbf{72.8} & \textbf{82.5} & \textbf{70.0}
        & \textbf{77.8} & \underline{65.9} & \textbf{84.6} & \textbf{63.2} & \underline{73.7} & \textbf{75.0} \\
        \bottomrule
    \end{tabular}%
    }
\end{table*}

\subsection{Reasoning-Type Analysis on EgoMemReason}
\label{sec:Reasoning-Type Analysis on EgoMemReason}

Table~\ref{tab:appendix_egomemreason_category} reports answer accuracy
by reasoning type on EgoMemReason. VMSA achieves 38.6\% overall accuracy
with Qwen3.6-27B and 41.4\% with GPT-5. Both VMSA configurations
show their largest gains on Event Ordering and Activity Pattern. With
GPT-5, VMSA reaches 60.0\% on Event Ordering and 58.0\% on Activity
Pattern, exceeding the best baseline in each category by 16.0 and
10.0\%, respectively. It also achieves the highest
accuracy on Cumulative State Tracking. The largest gaps relative to the best category-specific baselines remain
in Temporal Counting and Spatial Preference. 

\begin{table*}[!tbp]
  \centering
  \caption{Category-wise accuracy (\%) on the EgoMemReason benchmark~\cite{egomemreason}. The best and
           second-best results are shown in \textbf{bold} and \underline{underlined}, respectively.}
  \label{tab:appendix_egomemreason_category}
  \footnotesize
  \setlength{\tabcolsep}{3pt}
  \resizebox{0.85\textwidth}{!}{%
  \begin{tabular}{lccccccc}
    \toprule
    Method & Tracking & Counting & Ordering & Linking & Spatial & Activity & Overall \\
    \midrule
    \multicolumn{8}{l}{\textit{Base Models}} \\
    Qwen3-VL-8B~\cite{qwen3vl}        & 35.0 & 28.0 & 23.0 & 21.0 & \underline{40.0} & 42.0 & 29.6 \\
    GPT-5~\cite{gpt5}                 & 29.0 & 42.0 & 20.0 & 18.0 & 32.0 & 28.0 & 27.8 \\
    Gemini-3.1-Pro~\cite{gemini31pro} & \underline{40.0} & 26.0 & 44.0 & \textbf{33.0} & \underline{40.0} & 48.0 & 37.4 \\
    Qwen3.6-27B~\cite{qwen36}         & 33.0 & 29.0 & 35.0 & 25.0 & 32.0 & 30.0 & 30.6 \\
    \midrule
    \multicolumn{8}{l}{\textit{Long-Video LLMs}} \\
    LongVA-7B~\cite{longva}           & 22.0 & 18.0 & 20.0 & 22.0 & 20.0 & 22.0 & 20.6 \\
    StreamingVLM~\cite{streamingvlm}  & 25.0 & 29.0 & 21.0 & 20.0 & 20.0 & 32.0 & 24.2 \\
    InternVideo-2.5-8B~\cite{internvideo25} & 29.0 & 27.0 & 25.0 & 15.0 & 32.0 & 32.0 & 25.6 \\
    VideoLLaMA3-8B~\cite{videollama3} & 23.0 & 31.0 & 27.0 & \underline{32.0} & 38.0 & 36.0 & 30.0 \\
    Molmo2-8B~\cite{molmo2}           & 36.0 & \textbf{50.0} & 27.0 & 25.0 & 34.0 & 22.0 & 33.2 \\
    \midrule
    \multicolumn{8}{l}{\textit{Agentic Video Frameworks}} \\
    SiLVR~\cite{silvr} (GPT-5)        & 31.0 & 14.0 & 27.0 & 17.0 & 18.0 & 28.0 & 22.4 \\
    Ego-R1~\cite{egor1} (Qwen2.5-3B)  & 30.0 & 18.0 & 23.0 & 18.0 & \textbf{48.0} & 32.0 & 25.8 \\
    WorldMM~\cite{worldmm} (GPT-5)    & 32.0 & \underline{44.0} & 21.0 & 21.0 & 34.0 & 36.0 & 30.6 \\
    AVP~\cite{avp} (Gemini-2.5-Flash) & 34.0 & 42.0 & 31.0 & 27.0 & 38.0 & 34.0 & 34.0 \\
    \midrule
    \textbf{VMSA (Qwen3.6-27B)} & 37.0 & 28.0 & \underline{55.0} & 31.0 & 32.0 & \underline{52.0} & \underline{38.6} \\
    \textbf{VMSA (GPT-5)} & \textbf{41.0} & 29.0 & \textbf{60.0} & 31.0 & 34.0 & \textbf{58.0} & \textbf{41.4} \\
    \bottomrule
  \end{tabular}%
  }
  \par\vspace{2pt} 
  \parbox{0.85\textwidth}{\scriptsize
  \textbf{Task abbreviations:} Tracking: Cumulative State Tracking; Counting: Temporal Counting; Ordering: Event Ordering; Linking: Event Linking; Spatial: Spatial Preference; Activity: Activity Pattern.}
\end{table*}

\begin{table*}[!tbp]
  \centering
  \caption{Comparison on ScaleLong at the clip, shot, event, and story levels. We report accuracy (\%). The VMSA result uses memory skills transferred from other source benchmarks.}
  \label{tab:appendix_scalelong}
  \small
  \setlength{\tabcolsep}{3.5pt}
  \begin{tabular*}{0.75\textwidth}{@{\extracolsep{\fill}}lccccc@{}}
    \toprule
    Model / Method & Clip & Shot & Event & Story & Overall \\
    \midrule
    Gemini-2.5-Pro                           & 71.5 & 62.8 & 68.0 & 69.0 & 67.9 \\
    Doubao-1.5-VL-Pro                        & 66.4 & 52.8 & 55.2 & 60.2 & 58.7 \\
    InternVL2.5-78B~\cite{chen2024expanding} & 65.2 & 54.3 & 53.4 & 61.5 & 58.6 \\
    GPT-4o                                   & 61.8 & 50.7 & 51.0 & 58.0 & 55.4 \\
    Qwen3.6-27B                              & 65.1 & 55.9 & 52.1 & 58.6 & 58.0 \\
    VMSA (Qwen3.6-27B)                       & 79.0 & 62.5 & 67.1 & 67.6 & \textbf{69.1} \\
    \bottomrule
  \end{tabular*}
\end{table*}

%
\begin{table*}[!tbp]
  \centering
  \caption{Task-level comparison on the visual-only VideoOdyssey-V benchmark.
  We report accuracy (\%) for six Perception tasks, six Cognition tasks,
  Summarization, and Temporal Grounding. The VMSA result uses memory skills transferred from other source benchmarks.}
  \label{tab:appendix_videoodyssey}
  \setlength{\tabcolsep}{2.2pt}
  \resizebox{\textwidth}{!}{%
  \begin{tabular}{lrrrrrrrrrrrrrrr}
    \toprule
    & \multicolumn{6}{c}{Perception} & \multicolumn{6}{c}{Cognition}
      & \multicolumn{2}{c}{Other} & \\
    \cmidrule(lr){2-7}\cmidrule(lr){8-13}\cmidrule(lr){14-15}
    Model / Method & Count & ObRec & AcRec & AtRec & OCR & Cap
      & CaRea & EmRea & InRea & ObRea & SpRea & Order
      & Sum & TeGro & Overall \\
    \midrule
    \multicolumn{1}{l}{\# task labels / questions}
      & 307 & 208 & 192 & 106 & 167 & 49
      & 128 & 71 & 66 & 173 & 61 & 89
      & 65 & 65 & 1,618 \\
    \midrule
    GPT-5.2~\cite{gpt52}           & 28.3 & 55.3 & 46.9 & 51.9 & 64.1 & 49.0 & 51.6 & 43.7 & 60.6 & 60.1 & 45.9 & 69.7 & 44.6 & 44.6 & 49.0 \\
    Gemini-2.5-Pro                 & 31.6 & 59.6 & 49.5 & 54.7 & 60.5 & 44.9 & 52.3 & 49.3 & 62.1 & 60.1 & 41.0 & 64.0 & 61.5 & 40.0 & 50.4 \\
    Gemini-3.1-Pro                 & 34.2 & 63.0 & 57.8 & 64.2 & 66.5 & 59.2 & 59.4 & 45.1 & 78.8 & 65.3 & 50.8 & 70.8 & 56.9 & 49.2 & 56.3 \\
    Kimi-K2.5~\cite{team2026kimi}  & 26.7 & 57.7 & 50.0 & 51.9 & 52.7 & 44.9 & 57.0 & 52.1 & 62.1 & 53.8 & 39.3 & 57.3 & 69.2 & 46.2 & 48.6 \\
    Qwen3.5-27B                    & 30.9 & 43.8 & 50.0 & 59.4 & 48.5 & 51.0 & 48.4 & 42.3 & 50.0 & 52.6 & 41.0 & 55.1 & 49.2 & 21.5 & 44.6 \\
    Qwen3.6-27B                    & 34.5 & 48.6 & 48.4 & 46.2 & 53.3 & 55.1 & 50.8 & 53.5 & 48.5 & 51.4 & 39.3 & 40.4 & 50.8 & 40.0 & 45.9 \\
    VMSA (Qwen3.6-27B)
      & 44.3 & 78.4 & 70.8 & 74.5 & 79.0 & 69.4
      & 58.6 & 69.0 & 69.7 & 72.3 & 41.0 & 76.4
      & 67.7 & 80.0 & \textbf{66.4} \\
    \bottomrule
  \end{tabular}%
  }
  \vspace{2pt}

  \parbox{\textwidth}{\scriptsize
  \textbf{Task counts:} Task categories are not mutually exclusive; the
  1,618 unique questions have 1,747 task labels in total.\par
  \smallskip
  \textbf{Task abbreviations:} Count: Counting; ObRec: Object Recognition;
  AcRec: Action Recognition; AtRec: Attribute Recognition; Cap: Captioning;
  CaRea: Causal Reasoning; EmRea: Emotional Reasoning; InRea: Intentional
  Reasoning; ObRea: Object Reasoning; SpRea: Spatial Reasoning; Order: Temporal
  Ordering; Sum: Summarization; TeGro: Temporal Grounding.}
\end{table*}

\subsection{Results on ScaleLong and VideoOdyssey-V}

ScaleLong and VideoOdyssey-V require integrating evidence across multiple
temporal scales. Their videos are longer on average than those in the four
long-video benchmarks in the main comparison. For both target benchmarks,
we use the 19-skill library induced from four separate source benchmarks
(Section~\ref{sec:appendix-skill-induction}). On ScaleLong, VMSA achieves
69.1\% accuracy, exceeding the Qwen3.6-27B direct-video baseline by
11.1\% (Table~\ref{tab:appendix_scalelong}). On
VideoOdyssey-V, VMSA achieves 66.4\% accuracy, a gain of 20.5 points
over the same baseline (Table~\ref{tab:appendix_videoodyssey}).

\subsection{Effect of Subtitle Assistance}

Table~\ref{tab:appendix_with_subtitle} compares VMSA with and without
subtitles while holding the structured video memory configuration fixed.
Adding subtitles improves answer accuracy by 5.7\% on
Video-MME (L) and 2.5\% on LVBench, suggesting that subtitles
complement the visual evidence on both benchmarks.

\begin{table}[H]
  \centering
  \caption{Effect of subtitle assistance. Adding subtitles improves VMSA by
           5.7 and 2.5 points on Video-MME (L) and LVBench,
           respectively.}
  \label{tab:appendix_with_subtitle}
  \small
  \setlength{\tabcolsep}{5pt}
  \begin{tabular}{lcc}
    \toprule
    Dataset & No Sub. & w/ Sub. \\
    \midrule
    Video-MME (L) & 76.1 & 81.8 \\
    LVBench       & 72.8 & 75.3 \\
    \bottomrule
  \end{tabular}
\end{table}

\section{Computational Cost}

\subsection{Offline Memory-Construction Cost}
\label{sec:appendix-offline-cost}

Table~\ref{tab:appendix_costs_per_granularity} quantifies offline memory
construction beyond segment captioning on a matched subset of 10
LongVideoBench videos with an average duration of 22.30 minutes per video.
All measurements use Qwen3.5-9B on a single
NVIDIA RTX PRO 6000 and report per-video averages. At the default 16-second
granularity, memory construction uses 286{,}104 input tokens and 60{,}072
output tokens and takes 6.31 minutes per video. Increasing the segment
duration to 60 seconds reduces these costs to 77{,}229 input tokens, 13{,}667
output tokens, and 2.34 minutes per video. At the highest-accuracy 16-second
setting, the additional entity-centric memory is constructed in
$0.28\times$ the average video duration. Increasing segment duration further
reduces offline construction cost, at the expense of temporal resolution and
accuracy.

\begin{table}[htbp]
  \centering
  \caption{Average input tokens, output tokens, and memory-construction
           latency in seconds per video at different segment granularities,
           excluding segment captioning.}
  \label{tab:appendix_costs_per_granularity}
  \small
  \setlength{\tabcolsep}{5pt}
  \begin{tabular}{crrr}
    \toprule
    \shortstack{Segment\\duration} &
    \shortstack{Input\\tokens} &
    \shortstack{Output\\tokens} & \shortstack{Latency\\(s/video)} \\
    \midrule
    16 s (default) & 286{,}104 & 60{,}072 & 378.51 \\
    30 s           & 121{,}526 & 28{,}093 & 243.87 \\
    60 s           & 77{,}229  & 13{,}667 & 140.31 \\
    \bottomrule
  \end{tabular}
\end{table}

\subsection{Online Inference Cost}
\label{sec:appendix-online-cost}

Table~\ref{tab:appendix_inference_cost} summarizes the online cost of the
Qwen3.6-27B configuration of VMSA. Across the evaluated datasets, the agent
generates 1{,}303-3{,}755 agent tokens per question. Memory tokens comprise the initial Entity-Aware Episode
Memory, retrieved memory skills, and observations returned by Caption Search
and Entity Event Search. The average count ranges from 6,501 on HippoVlog to
45,795 on EgoMemReason, reflecting the larger amount of memory content used
on week-long egocentric benchmarks. On the other long-video benchmarks,
Segment Inspection decodes from 99.9 to 151.3 frames per question on average.


\begin{table}[htbp]
  \centering
  \caption{Average online inference cost per question for the Qwen3.6-27B
           agent. \emph{Tool turns} count interaction rounds; \emph{agent
           tokens} count generated assistant tokens;
           \emph{memory tokens} count the initial Entity-Aware Episode Memory,
           retrieved memory skills, and Caption Search and Entity Event Search observations; and \emph{visual
           frames} count frames decoded for Segment Inspection. Across datasets, the agent
           averages from 2.07 to 4.59 tool turns per question and decodes at most
           151.3 frames per question on average.}
  \label{tab:appendix_inference_cost}
  \small
  \setlength{\tabcolsep}{2pt}
  \resizebox{\columnwidth}{!}{%
  \begin{tabular}{lrrrr}
    \toprule
    Dataset & \shortstack{Tool\\turns} & \shortstack{Agent\\tokens}
            & \shortstack{Memory\\tokens} & \shortstack{Visual\\frames} \\
    \midrule
    Video-MME (L)   & 3.90 & 1,847 & 14,864 & 108.1 \\
    LongVideoBench (L) & 4.59 & 1,592 &  7,804 & 149.7 \\
    LVBench         & 3.23 & 1,448 &  8,551 & 151.3 \\
    HippoVlog       & 2.60 & 1,303 &  6,501 &  99.9 \\
    EgoLifeQA       & 2.42 & 2,662 & 36,954 &   - \\
    EgoMemReason    & 2.07 & 3,755 & 45,795 &   - \\
    Ego-R1 Bench      & 2.35 & 1,607 & 38,072 &   - \\
    \bottomrule
  \end{tabular}
  }
\end{table}

\subsection{Effect of Search-Turn Budget}
\label{sec:abl_turn_budget}

We examine how the maximum search-turn budget affects VMSA's answer
accuracy. Table~\ref{tab:ablation_turn_budget} shows that increasing the
budget from 3 to 7 turns improves accuracy on all three benchmarks. The
gains saturate at 7 turns on LVBench and Video-MME~(L). On
LongVideoBench~(L), accuracy improves at every budget increase, rising
from 62.8\% at 3 turns to 71.1\% at 7 turns and 74.1\% at 10 turns.
VMSA also uses more turns on average on LongVideoBench~(L) than on any
other benchmark, which suggests that some questions
benefit from a larger budget to gather sufficient evidence. 

\begin{table}[htbp]
  \centering
  \caption{Effect of the maximum search-turn budget on answer accuracy (\%).
           The best results are shown in \textbf{bold}.}
  \label{tab:ablation_turn_budget}
  \setlength{\tabcolsep}{6pt}
  \resizebox{\columnwidth}{!}{%
  \begin{tabular}{c ccc}
    \toprule
    Max Turns & LongVideoBench (L) & LVBench & Video-MME (L) \\
    \midrule
    3            & 62.8 & 69.5 & 73.9 \\
    5            & 69.7 & 71.6 & 74.6 \\
    7            & 71.1 & \textbf{72.8} & \textbf{76.6} \\
    10 (default) & \textbf{74.1} & \textbf{72.8} & 76.1 \\
    \bottomrule
  \end{tabular}%
  }
\end{table}

\clearpage
\begin{figure*}[p]
  \centering
  \includegraphics[width=\textwidth,height=0.92\textheight,keepaspectratio]{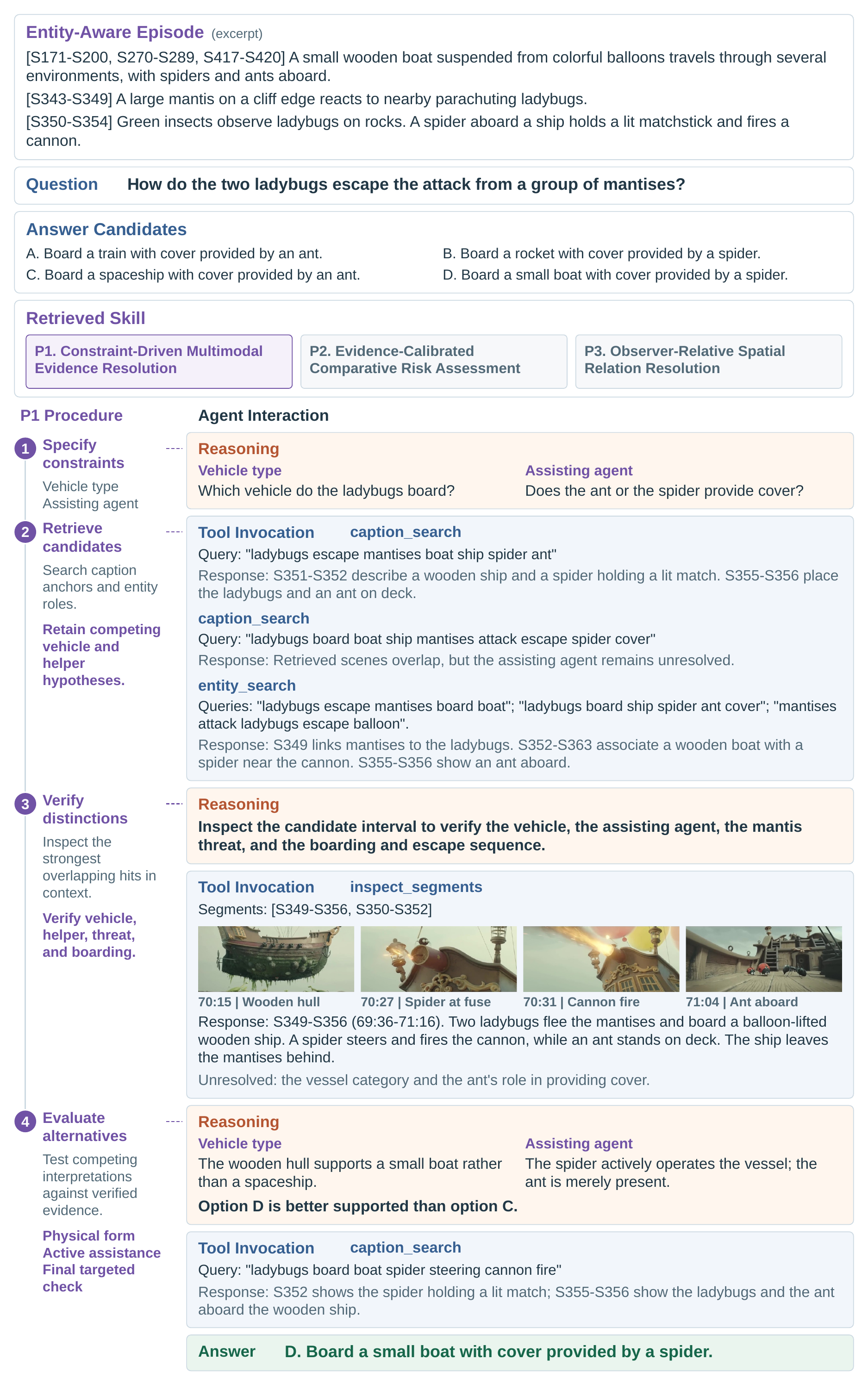}
  \caption{\textbf{Qualitative example of skill-guided agentic search on LVBench.}
  Retrieved memory skills guide Caption Search, Entity Event Search, and
  Segment Inspection to identify the escape vehicle and distinguish the
  spider's active assistance from the ant's presence aboard.
  Agent interactions and tool responses are summarized for clarity.}
  \label{fig:appendix-vmsa-example}
\end{figure*}

\end{document}